\documentclass[bst/sn-mathphys,Numbered]{sn-jnl}

\usepackage{geometry}

\newgeometry{vmargin={25mm}, hmargin={35mm,35mm}}   

\usepackage{graphicx}%
\usepackage{multirow}%
\usepackage{amsmath,amssymb,amsfonts}%
\usepackage{amsthm}%
\usepackage{bbold}%
\usepackage{mathrsfs}%
\UseRawInputEncoding
\usepackage[title]{appendix}%
\usepackage{xcolor}%
\usepackage{textcomp}%
\usepackage{manyfoot}%
\usepackage{booktabs}%
\usepackage{algorithm}%
\usepackage{xcolor}
\usepackage{listings}%
\usepackage{makecell}%
\usepackage{float}%
\usepackage{caption}%
\usepackage{subcaption}%
\usepackage{soul}%
\usepackage{algpseudocode}
\usepackage{bbm}
\usepackage{float}

\theoremstyle{thmstyleone}%
\theoremstyle{thmstyletwo}%

\theoremstyle{thmstylethree}%

\begin{document}

\title[Article Title]{Binary Classification: Is Boosting stronger than Bagging?}


\author*[1,2]{\fnm{Dimitris} \sur{Bertsimas}}\email{dbertsim@mit.edu}

\author[1]{\fnm{Vasiliki} \sur{Stoumpou}}

\affil*[1]{\orgdiv{Operations Research Center}, \orgname{Massachusetts Institute of Technology}, \orgaddress{\city{Cambridge}, \state{MA}, \country{USA}}}

\affil[2]{\orgdiv{Sloan School of Management}, \orgname{Massachusetts Institute of Technology}, \orgaddress{\city{Cambridge}, \state{MA}, \country{USA}}}

\abstract{Random Forests have been one of the most popular bagging methods in the past few decades, especially due to their success at handling tabular datasets. \textcolor{black} {They have been extensively studied and compared to boosting models, like XGBoost, which are generally considered more performant.} Random Forests adopt several simplistic assumptions, such that all samples and all trees that form the forest are equally important for building the final model. We introduce Enhanced Random Forests, which is an extension of vanilla Random Forests with extra functionalities and adaptive sample and model weighting. We develop an iterative algorithm for adapting the training sample weights, by favoring the hardest examples, and an approach for finding personalized tree weighting schemes for each new sample. Our method significantly improves upon regular Random Forests across 15 different binary classification datasets and considerably outperforms other tree methods, including XGBoost, when run with default hyperparameters, which indicates the robustness of our approach across datasets, without the need for extensive hyperparameter tuning. Our tree-weighting methodology \textcolor{black}{results in enhanced or comparable performance to the uniformly weighted ensemble}, and is, \textcolor{black}{more importantly}, leveraged to define importance scores for trees based on their contributions to classifying each new sample. This enables us to only focus on a small number of trees as the main models that define the outcome of a new sample and, thus, to partially recover interpretability, which is critically missing from both bagging and boosting methods. \textcolor{black} {In binary classification problems, the proposed extensions and the corresponding results suggest the equivalence of bagging and boosting methods in performance, and the edge of bagging in interpretability by leveraging a few learners of the ensemble, which is not an option in the less explainable boosting methods.}}




\maketitle

\section{Introduction}\label{sec:intro}
Decision tree methods have gained significant popularity in machine learning, largely due to their rapid training process and their clear interpretability. Each tree is composed of multiple splits, typically on individual features, making the classification process entirely transparent to the user. The most widely used tree algorithm, CART, introduced by \cite{breiman2017classification}, has seen extensive application; although the trees are trained based on heuristic methods, they can be grown very fast.

However, they are associated with multiple limitations. Although trees can learn non-linear functions, the nature of their splits make them successful when the function to learn is piecewise-constant, but they generally struggle with more complicated functions. Also, they are prone to both under- and overfitting \cite{bramer2007avoiding}; shallow trees only use a small subset of the available features, which might not successfully capture the complexity of the dataset, whereas deeper trees might have leaf nodes with very few training samples, and thus be very tailored to the training set distribution. In addition, small changes to the dataset can result in different splits and as a result a different model, which can significantly change its predictions, making trees unstable \cite{li2002instability}. 

The performance of trees can be significantly improved if they are ensembled, since then the limitations associated with a single tree are avoided. The ensemble reduces overfitting and results in a more stable model. There are two main families of ensemble learning algorithms that leverage trees as their learners: bagging and boosting methods. Both of them are collections of weak learners, but they utilize them in different ways. 

Bagging, introduced by \cite{breiman1996bagging} involves training multiple models independently from each other in parallel and then combining their predictions to determine the final output. Each model is trained on a subset of the dataset, selected with replacement (bootstrap sampling). The most popular example of this method is Random Forests \cite{breiman2001random}, where the learners combined are trees (CARTs). Random Forests have been shown to significantly improve the performance of single decision trees and mitigate the aforementioned issues related with them. 

Boosting is a different ensembling method, in which multiple weak learners are trained sequentially, in order to eventually build a more powerful model \cite{schapire1999brief}. More specifically, an initial model is trained and then a second model is trained on the residuals of the outputs and the actual ground truth labels of the first model, in order to correct the first model's errors. This process continues iteratively, and in the end the learners are combined to make the final prediction. There are multiple boosting approaches; the most popular has been XGBoost \cite{chen2016xgboost}, which implements a gradient boosting algorithm with very high efficiency. 

In practice, although both bagging and boosting methods both improve on single decision trees, it is usually experimentally shown that boosting has a performance edge over bagging; in fact XGBoost is considered one of the state-of-the-art methods for both classification and regression problems that involve tabular data. Albeit strong performing, bagging and boosting models lack interpretability, since the final prediction is now made by a collection of models, rather than a single tree.

\subsection{Contributions}

This work focuses on binary classification, which is one of the most common classification problems, and aims to extend the current Random Forest framework, to incorporate extra functionalities that make it more competitive against boosting methods. We refer to the new model as Enhanced Random Forest, since it is enriched with additions that control the trees' randomness in a targeted way, leveraging the training set to improve out-of-sample performance and interpretability. 

The first major contribution of this work is the enhancement of Random Forests' performance by introducing sample importance in the Random Forest algorithm; usually, in most machine learning problems, training samples are treated as equally important during the training of the model. In this work, we focus more on the hardest samples from the training set, and thus favor them with a higher weight; we develop an iterative algorithm that adaptively updates weights of the training samples and the probability of them being selected by the trees and we empirically show that this method significantly improves the Random Forest performance across 15 different datasets. \textcolor{black}{More specifically, sample importance weights result in an AUC performance increase in 13 out of 15 datasets (around 2\% on average, with a 7.5\% maximum improvement) and comparable performance to XGBoost. The difference between the two methods becomes statistically significant in favor of Enhanced Random Forests when we use default hyperparameters, indicating the robustness of the approach across datasets.}

The second contribution of our framework is the partial introduction of interpretability by assigning weights to the different learners too; namely, the final prediction for a new sample should not necessarily be the result of equivalent contribution of all the available models, since some of the learners might be more suitable for some of the out-of-sample points compared to others. We address this observation by assigning, for each sample, larger weights to models that correctly classify training points similar to the new sample. This approach facilitates the utilization of model weights as model importance proxies for each new sample separately, which enables users to locate a small collection of trees that contribute more to the final prediction for this specific sample, \textcolor{black}{while providing improved or similar performance to the uniform model weighting case.} By studying a handful of models, the user can get an insight into the main factors that play a role in the classification of a new sample, in a personalized way, thus partially recovering the interpretability missing from the regular Random Forest and XGBoost methods.

Overall, our framework provides a flexible extension of vanilla Random Forests, and incorporates the ideas of unequal sample and learner importance in a way that increases the out-of-sample performance to the levels of boosting algorithms and even outperforms them in a default parameter setting, without extensive hyperparameter tuning. It also enables interpretability recovery, which is essential in multiple real-life applications; this approach leverages the explainable learners that comprise the Random Forest, which is not the case in boosting methods, where individual trees are residual learners and not predictors for the classification problem. From these two perspectives, we demonstrate that in the case of binary classification bagging can be considered equally, if not more powerful, than boosting methods.  

\subsection{Related Work}\label{sec:related-work}

Sample weighting is a method that has been explored in the context of statistics and various machine learning algorithms. 

The idea of importance sampling has been thoroughly studied, starting with \cite{kahn1953methods}. Importance sampling is used to improve the efficiency of sampling in Monte-Carlo calculations and has been studied by \cite{Elvira2016HereticalMI, Paananen2019PushingTL}. Still in the Statistics field, \cite{barratt2021optimal} introduce optimal representative weighting, which aims to find weights so that sample averages are close to specific prescribed values, by formulating this as a (usually) convex optimization problem. 

In Machine Learning, typically, the focus of sample weighting is to mitigate the class imbalance problem often encountered in datasets. Namely, the goal is to assign larger weights to samples from the underrepresented class, usually proportionally to the number of samples per class. \cite{cui2019class} propose an alternative of adding class weights based on the volume of each point, which is calculated taking into consideration a small neighborhood around the point and is shown to improve performance in long-tailed and large-scaled datasets. 

Also, an approach that is often studied and inspired this work as well is assigning larger weights to harder samples. There are differentiations in the weighting schemes applied, as well as in the definition of hardness of a sample. The most popular algorithm that incorporated sample weights is AdaBoost \cite{freund1997decision}. AdaBoost is a boosting algorithm in which, at each step, before training a new learner, the model and sample weights are adjusted, so that the best learners and the samples misclassified by the previous learners (harder samples) get assigned a higher weight.

Curriculum \cite{bengio2009curriculum} and self-paced learning \cite{kumar2010self} refer to two families of algorithms that consider sample importance during training, by following the concept that the model should be first presented with the easiest samples and then gradually learn from the harder ones too. In curriculum learning, the easiness of the samples is known prior to training the model, whereas in self-paced learning it is determined by the confidence of the model predicting the samples' true outputs. Self-paced curriculum learning, introduced by \cite{jiang2015self}, combines the two approaches and incorporates prior knowledge about the samples in the dynamic self-paced learning framework. 

Sample weighting has also been extensively studied in Deep Learning. Focal Loss \cite{lin2017focal} down-weights the
loss assigned to well-classified examples and has been shown to improve dense object detection in computer vision. \cite{santiago2021low} propose an optimization framework to automatically determine how much each sample should contribute to the loss function. Recently, the idea of meta-learning in sample weighting has also emerged, with approaches like \cite{ren2018learning}, where the sample weights are determined by a meta gradient descent step on the current weights to minimize the loss on a hold-out validation set. Another method introduced by \cite{shu2019meta} aims to parameterize a weighting function for the samples using an MLP and learn it simultaneously with training the classification model.

Focusing on extensions of Random Forests, there have been multiple works exploring sample, feature and model weights. \cite{maudes2012random} construct RFWs (Random Feature Weights), which are similar to Random Forests, with one difference being that in each node of the trees, all samples are considered, and a few are randomly selected with a non-uniform probability distribution, unlike Random Forests, where this probability is actually uniform. This distribution is different per tree. In a similar fashion, \cite{liu2017variable} propose variable importance-weighted Random Forests, by again sampling features according to their feature importance scores. 

More closely related to our proposed framework, \cite{kim2011weight} introduce a weighted voting classification ensemble method (WAVE) that iteratively determines weights for samples and models. Samples that are harder to classify are assigned a higher weight, and models that are better at correctly classifying the hardest samples also get a higher weight. Although they report their results on multiple datasets, the number of samples and trees they consider are relatively small (up to 64 models in the ensemble), and the difference in performance between WAVE and Random Forests is minor as the number of models increases.

\cite{robnik2004improving} follows an approach of model weighting which bears similarities to our approach; for each new sample, the closest training points are found by calculating a similarity score based on how often the samples end up in the same leaf node. Then, the trees that are better performing (in terms of a margin function they define) at these training points are considered to make the final prediction for the new sample. Another tree-weighting approach is introduced by \cite{li2010trees}, who select the accuracy of each tree in out-of-bag data as the indication of the tree's performance and set it as the tree's weight.   \cite{winham2013weighted} develop Weighted Random Forests, which use tree-level weights to favor trees with higher prediction accuracy in out-of-bag samples. However, out-of-bag samples are not the same for each tree, which means that defining weights this way is not necessarily fair. To mitigate this issue, \cite{xuan2018refined} develops Refined Weighted Random Forests, where the weights are determined by both in- and out-of-bag data. \cite{shahhosseini2021improved} experiment with different ways of finding tree weights, by trying to maximize accuracy, AUC or performing stacking of the models. Another approach is introduced by \cite{chen2023optimal}, who propose Mallow's-Type Weighted RFs, by treating the individual trees within the RF as base learners and employing Mallows-type criteria to obtain their respective weights.

Our approach focuses both on sample and model weights; sample weights generally result in a significant performance boost compared to the regular Random Forests, and model weighting is also beneficial in most datasets, especially when applied on vanilla Random Forests directly. The main difference with existing tree-weighting approaches lies on the fact that we leverage the resulting weights to increase the interpretability of the model. 

\textcolor{black}{Interpretability in Random Forests and boosting methods is mostly explored through feature importance scores; each feature is assigned a score based on how much it contributes in the final prediction. The most commonly used feature importance score is the mean decrease in impurity (MDI), but other metrics are also employed. An issue with feature importance scores is that they do not indicate how the value of the feature actually affects the model prediction. To address this, Shapley values (\cite{lundberg2017unified}) are often used. They provide an insight into how much and in which way the value of each feature affects the prediction of the model; however Shapley values are generally complex to compute. Our approach leverages the interpretable nature of the trained learners and focuses on the ones that are more important for each new sample individually, providing a personalized and in-depth approximate interpretability feature, without extra computational burden.}

\section{Methods}

In this section, we describe the algorithms used to implement all the additional functionalities added in our proposed extension of Random Forests. Section \ref{subsec:vanilla_rf} describes in more detail how regular Random Forests work, while Sections \ref{subsec:enhanced_rf} and \ref{subsec:interpretability} discuss our framework of Enhanced Random Forests and its interpretability dimension respectively .  

\subsection{Vanilla Random Forests} \label{subsec:vanilla_rf} 

Random Forests (\cite{breiman2001random}) are collections of simple Decision Trees (CARTs), where each tree is trained on a random subset of samples and features of the training set. The randomness in sample and feature selection in each tree results in the creation of different trees, that can be used as different predictors in an ensemble learning setting. This has empirically shown to considerably improve the performance of a single CART, by sacrificing, however, interpretability, since the final result is given by a large collection of models. 

More specifically, each tree is trained on a random subset of training samples (rows), selected with replacement, a method called bootstrap sampling. While building each tree, the best split variable must be chosen for each node; namely, for each node, typically all possible features (variables) are considered, and the one that gives the split that results in the highest score (e.g. Gini impurity) is selected. In Random Forests, for each split in each Decision Tree, only a random subset of features is considered, further increasing the element of randomness in the training of the model. The best split is chosen from the selected features based on a criterion like Gini impurity (for classification). 

The resulting ensemble is used to make the predictions. In classification problems, the probability of each sample $i$ belonging to each class is obtained from each tree separately, and then the average probability $p_i$ is calculated.

\subsection{Enhanced Random Forests} \label{subsec:enhanced_rf}

An assumption consistently made when training and using a Random Forest, is that all samples are equally important; thus, sample weights are all equal to one. Also, when calculating the final prediction, each tree contributes equally to the calculation of the average probability. These approaches are simplistic, and hide the fact that some samples might be more important than others when training the model, and also that each tree should not necessarily contribute equivalently to calculating the prediction for each sample. 

To address these observations, we develop an approach that aims to extend the typical Random Forest framework to facilitate more functionalities. More specifically, our framework consists of two major extensions: sample and feature handling, and model weighting.


The first extension mostly focuses on treating the samples based on their custom importance, which is determined by their difficulty. Sample importance can be incorporated in the model's training algorithm by assigning weights to samples, which are used while building the trees, or by assigning a different probability for each sample, so that when the samples are randomly selected by each tree, the selection distribution is non-uniform. The combination of the two methods is also shown to be beneficial. We propose an iterative approach that defines a customized weight for each sample, and this weight is used as a sample weight in the training process, as well as to define a selection probability for this sample. In each iteration, a Random Forest is trained, and the weight is calculated for each training sample by utilizing its misclassification error by the model, so that harder samples are favored. By using this calculation of sample importance, and by calculating the feature importance using the mean decrease in impurity (MDI), the dataset can be iteratively cleaned, by discarding the least important features and samples.

The second extension focuses on model weighting, which is performed in a personalized way, so that each new sample has a specific tree weighting scheme associated with it. For each sample, its nearest neighbors from the training set are found and only the trees that perform better on them are used to make the prediction for the sample. Depending on the number of neighbors that are considered and the number of highest-ranked models, fewer models determine the output for the sample, and each of them contributes to the final prediction with a different weight. Using this weight and the performance of the models on the training set neighbors, a score for each tree is defined, and the user can inspect the highest-scoring trees to get an insight into the factors that led to the forest making the final decision for the sample. This enables partial interpretability recovery and can be useful in multiple real life applications. 

The following sections thoroughly illustrate the two aforementioned extensions. 

\subsubsection{Sample \& Feature importance} \label{subsec:sample_weighting}

As already mentioned, typically in most machine learning problems, all training samples are treated equally, namely there is no distinction between samples in the training process. The way with which sample importance affects the training process depends on the machine learning model employed to solve the problem. In the case of Random Forests, we can assign weights to the training samples, so that each training sample is associated with a specific weight. These weights affect the scores (e.g. Gini impurity score) computed at each node of the trees when the splits are determined, and can thus play a significant role in the final structure of each tree. In our approach, we aim to assign larger weights to the hardest samples, namely the ones that are misclassified by the model. Additionally, since each tree in the Random Forest algorithm is trained using a random subset of the training samples, we differentiate the importance of the samples by assigning a different selection probability for each sample. To favor the hardest training samples, a higher probability is assigned to them, in the same fashion as sample weights are assigned.  

Formally, we consider a dataset with $N$ samples, with features $\mathbf{X}$ and targets $\boldsymbol{y}$. For the purpose of this work, we focus on binary classification problems, so for each target $y_i$, $i=1, \dots, N$, $y_i \in \{0,1 \}$. We split the dataset into a training, validation and test set, denoted as: $\mathbf{X}_{train}, \boldsymbol{y}_{train}$, $\mathbf{X}_{val}, \boldsymbol{y}_{val}$ and $\mathbf{X}_{test}, \boldsymbol{y}_{test}$ respectively. 

Each sample $i=1, \dots, N$ of the training set is associated with a specific weight $w_i \in \mathbb{R}$.  The outline of our approach is as follows:

\begin{itemize}
    \item Equal weights $w_i = 1$, $\forall i=1, \dots, N$, and equal probabilities $s_i =  \frac{w_i}{\sum_{i=1}^N w_i}$ are initialized. 
    \item A Random Forest is trained on the training set ($\mathbf{X}_{train}$, $\boldsymbol{y}_{train}$), using sample weights $w_i$ and sample probabilities $s_i$, $i=1, \dots, N$.
    \item Then, the classification threshold $t$ is calculated, which serves as the boundary between the two classes. This means that, for each sample $i$, the model provides the probability $p_i$ of $i$ being classified as 1, and if $p_i \geq t$, then sample $i$ is classified as 1, and as 0 otherwise. To estimate $t$, the true positive (TPR) and false positive rates (FPR) for different threshold $t$ values are first calculated. The threshold $t$ is selected using Youden's J statistic, which chooses the $t$ for which the difference (TPR-FPR) is maximized:
    \begin{equation}
        t^* = \text{argmax}_{t} (TPR(t) - FPR(t)).
    \end{equation}
    \item The next step is to update the weights, according to the equation:
    \[
        w_i' = 
        \begin{cases} 
        w_i + \lambda (t^*-p_i), & \text{if } y_i = 1, \\
        w_i + \lambda (p_i - t^*), & \text{if } y_i = 0, 
        \end{cases}
    \]
    where $\lambda$ is a learning rate hyperparameter. The purpose of this assignment is to give larger weights to misclassified samples, proportionally to their misclassification error. For example, a sample from class 1 that has $p_i <t^*$ is misclassified, and the highest the misclassification error $t^*-p_i$, the highest the weight increase for this sample. On the contrary, if $p_i$ is considerably larger than $t^*$, the sample is considered easy, and thus the weight assigned to it is reduced. A similar argument holds for the case where $y_i = 0$. Negative weights are counterintuitive; in case $w_i <0$, we set $w_i = 0$. The selection probabilities are also updated as $s_i =  \frac{w_i}{\sum_{i=1}^N w_i}$.
    \item This process (training a Random Forest and updating the sample weights and probabilities) continues until a stopping criterion is met. For example, one possible option is to decide a specific number of iterations beforehand and select the best model based on the performance on the validation set ($\mathbf{X}_{\text{val}}$, $\boldsymbol{y}_{\text{val}}$), or to employ an early stopping approach and interrupt the process when the validation set performance starts dropping. 
\end{itemize}

The full algorithm is described in short in Algorithm \ref{alg:training_with_weights}.

\begin{algorithm}
\caption{Enhanced Random Forest Algorithm}
\begin{algorithmic}[1]
\State Initialize weights $w_i = 1$, $\forall i=1, \dots, N$.
\State Initialize probabilities $s_i =  \frac{w_i}{\sum_{i=1}^N w_i}$.
\Repeat
    \State Train a Random Forest on the training set ($\mathbf{X}_{\text{train}}$, $\boldsymbol{y}_{\text{train}}$) using sample weights $w_i$ and probabilities $s_i$, $\forall i=1, \dots, N$.
    \State Calculate the classification threshold $t$:
    \begin{enumerate}
        \item For each sample $i$, compute the probability $p_i$ of $i$ being classified as 1.
        \item Calculate the ROC curve using the training set labels, obtaining TPR and FPR for different $t$ values.
        \item Select the threshold $t$ that maximizes Youden's J statistic: $J = (\text{TPR} - \text{FPR})$.
    \end{enumerate}
    \State Update the weights $w_i$:
    \[
        w_i' = 
        \begin{cases} 
        w_i + \lambda (t - p_i), & \text{if } y_i = 1, \\
        w_i + \lambda (p_i - t), & \text{if } y_i = 0. 
        \end{cases}
    \]
    \State If $w_i <0$, we set $w_i=0$.
    \State Update probabilities $s_i =  \frac{w_i}{\sum_{i=1}^N w_i}$.
\Until{stopping criterion is met. Possible stopping criteria are:
    \begin{itemize}
        \item Predefined number of iterations.
        \item Early stopping based on validation set performance ($\mathbf{X}_{\text{val}}$, $\boldsymbol{y}_{\text{val}}$).
    \end{itemize}}
\end{algorithmic}  \label{alg:training_with_weights}
\end{algorithm}




\paragraph{Selection of Features} \label{subsec:sample_feature_cleaning}
As described in Section \ref{subsec:vanilla_rf}, a random subset of features is selected at each split of each tree, and the one that results in the highest score is eventually chosen. In our framework, we facilitate the feature selection to happen for each tree, rather than each split of the tree. So, each tree only takes into account a random subset of the features, and these are considered to find the best split as the tree grows. We find that this extension is helpful in some of the datasets, highlighting the potential improvement it can offer in some cases, as presented in Section \ref{subsec:effect_of_ext}.

\paragraph{Sample \& Feature cleaning}
The concept of assigning weights or different probabilities to samples is, as already mentioned, closely related to the notion of sample importance, since samples with very low weights or probabilities contribute less to building the final model and are thus considered less important. This observation can be leveraged to automatically clean the training set and remove samples that do not have an actual effect on the creation of the model, or might even hurt its performance. More specifically, the weights/probabilities assigned to the samples can serve as importance scores, where clearly the lower the value, the lower the importance. The samples with low importance values can be discarded. 

In terms of feature cleaning, feature importance scores are utilized, to again discard features with low importance. We employ the mean decrease in impurity (MDI) as the feature importance score, since this is used by the main commercial available packages. For a given node \( t \), let \( I(t) \) be the impurity of the node before splitting, and \( I_L(t) \) and \( I_R(t) \) be the impurities of the left and right child nodes after the split. The MDI of feature \( j \) for this split is:

\[
\Delta I_j(t) = I(t) - \left( \frac{n_L}{n} I_L(t) + \frac{n_R}{n} I_R(t) \right),
\]

\noindent where \( n \) is the number of samples in node \( t \), \( n_L \) and \( n_R \) are the number of samples in the left and right child nodes, respectively. The importance of feature \( j \) for a single tree is the sum of \( \Delta I_j(t) \) over all nodes \( t \) where \( j \) is used for splitting. For the entire forest, the importance of feature \( j \) is the average importance over all trees.

Using these two scores, features and samples are removed in an iterative fashion, until a noticeable performance deterioration is observed in a hold-out validation set. We adopt an approach where the sample importance scores are sorted, grouped in buckets and then the bucket with the lowest scores is discarded. When a feature importance score is lower than a proportion of the highest score across the features, the feature is discarded. The different approach followed for samples and features is attributed to the fact that features are usually less than samples, and the sample cleaning approach is more aggressive. The approach is described in more detail in Algorithm \ref{alg:sample_feature_cleaning}. The value of $l$ is selected based on how quickly the user wants to discard features. We typically set $l=0.1$.

There are multiple criteria that can be chosen in terms of the way of discarding samples, and our approach is one of the reasonable ways that the sample and feature cleaning can be performed. For example, one can select a threshold and discard samples or features whose importance score is below this threshold. However, determining a specific threshold can be restrictive, since the importance scores can vary for different datasets, whereas our approach does not need tuning for each individual dataset. 

\begin{algorithm} 
\caption{Training Algorithm with Sample and Feature Cleaning}
\begin{algorithmic}[1]
\Repeat
    \State Train an Enhanced Random Forest.
    \If {sample\_cleaning}
        \State Sort the samples in descending order based on importance score \( s_i \).
        \State Create buckets of samples.
        \State Discard the bucket with the lowest scores.
    \EndIf
    \If {feature\_cleaning}
        \State Sort the features in descending order based on importance score \( f_j \).
        \State Discard features \( j \) for which \( f_j < l \cdot f_{(1)} \), where \( f_{(1)} \) is the highest feature importance.
    \EndIf
\Until{stopping criterion is met. Possible stopping criteria are:
    \begin{itemize}
        \item Predefined number of iterations.
        \item Early stopping based on validation set performance.
    \end{itemize}}
\end{algorithmic} \label{alg:sample_feature_cleaning}
\end{algorithm}

\subsubsection{Model weighting} \label{subsec:model_weighting}
As described in Section \ref{subsec:vanilla_rf}, the final prediction for each sample is determined by the average probability across all trees. More specifically, suppose that we have a forest $\mathcal{F}$, that contains $N_t$ trees $\mathcal{T}_i$, $i=1, \dots N_t$. For a sample $k$, the probability of $k$ belonging to class 1 is calculated as:
\begin{equation}
    p_k = \frac{1}{N_t} \sum_{i=1}^{N_t} p_{i, {\mathcal{T}_i}},
\end{equation}
where $p_{i,{\mathcal{T}_i}}$ is the probability of sample $k$ belonging to class 1, as calculated from tree $\mathcal{T}_i$. However, it is reasonable that not all trees are equally important for each sample; some of the trees of the ensemble might completely misclassify a sample, especially in a Random Forest setting, where the randomness in sample and feature selection results in overall weaker learners. The goal is to calculate the final probability of sample $k$ as the weighted average of the trees' probabilities:
\begin{equation}
    p_k = \frac{1}{N_t} \sum_{i=1}^{N_t} u_{k,i} p_{i, {\mathcal{T}_i}}.
\end{equation}

To select the model weights $u_{k,i}$, $i=1, \dots ,N_t$, we develop an approach which assigns a different tree weight combination for each new data point, creating a personalized prediction. The goal is to favor trees that are more accurate for a specific point. This accuracy needs to be approximated, since the ground-truth label for a test set sample is not known. To achieve that, for each point, the trees that make the best predictions for its nearest neighbors in the training set are found, and only these are used to calculate the final probability for this point. 

More formally, we consider a training set of $N$ points. For each sample in the training set, the "best" trees are found; namely, the trees, whose output probability for this sample is closest at predicting its ground-truth class. For example, in our binary classification case, suppose we have a training point $\boldsymbol{z}$. If the class of point $\boldsymbol{z}$ is $y_{\boldsymbol{z}} =1$, the trees are ranked in a descending order $\tau_{\boldsymbol{z},(1)}, \tau_{\boldsymbol{z},(2)}, \dots, \tau_{\boldsymbol{z},(N_t)}$, so that $p_{i,\tau_{\boldsymbol{z},(1)}} \geq p_{i,\tau_{\boldsymbol{z},(2)}} \geq \dots \geq p_{i,\tau_{\boldsymbol{z},(N_t)}}$. The order is ascending in case $y_{\boldsymbol{z}}=0$. So, the training point $\boldsymbol{z}$ has a collection of trees $\tau_{\boldsymbol{z},(i)}, i=1, \dots, N_t$ associated with it with a specific order, which is unique for this point, with $\tau_{\boldsymbol{z},(1)}$ being the highest ranked tree for point $\boldsymbol{z}$. 

Suppose we have an out-of-sample point $k$. In order to specify the weights $u_{k,i}$, $i=1, \dots ,N_t$, its nearest neighbors from the training set are found, using the Euclidean distance, and only the best trees of these neighbors are considered. Assuming that its $M$ closest training points (nearest neighbors), which we denote as the set $\mathcal{M}_k$, are detected and the $L$ best trees of each point are taken into account, the probability for the new sample $k$ is: 

\begin{equation}
    p_k = \frac{1}{M \cdot L} \sum_{\boldsymbol{m} \in \mathcal{M}_k} \sum_{l=1}^{L} p_{k,\tau_{\boldsymbol{m},(l)}}.
\end{equation}

This means that the weights $u_{k,i}$, $i=1, \dots, N_t$ are defined as:

\begin{equation}
    u_{k,i} = \frac{1}{M \cdot L} \sum_{\boldsymbol{m} \in \mathcal{M}_k} \sum_{l=1}^{L} \mathbb{1} \{ \tau_{\boldsymbol{m}, (l)}  = \mathcal{T}_i\}
\end{equation}

\noindent for each point $k$. In this way, trees that do not give good predictions for points that are close to $k$ can even have zero weight in the calculation, whereas points that appear often as the "best" for close neighbors get large weights. The hyperaparameters of this approach is the number of the nearest neighbors from the training set ($M$) and the number of trees that are considered for each neighbor ($L$). These can be selected based on the performance on a hold-out validation set. 

\subsection{Interpretability} \label{subsec:interpretability}

The idea of personalized tree importance for each sample individually can be leveraged to increase the interpretability of Enhanced Random Forests. More specifically, being an ensemble model consisting of multiple trees, Random Forests generally lack explainability, since the final prediction is the average prediction of multiple trees. The model weighting method described in Section \ref{subsec:model_weighting} produces weights $u_{k,i}$, which specify which of the trees contribute more in the calculation of the final probability for each point. So, by focusing on the trees with the largest weight for each sample, a handful of interpretable models mostly contribute in the final prediction, and thus offer an insight into the features that led to this classification. Clearly, studying a few trees is feasible, compared to hundreds of equally weighted trees, as is the case for typical Random Forests. This idea is unique to the family of random forest algorithms, since boosting methods include trees that are trained on the residuals of the predictions and are, thus, not intuitive for the users.

The weights $u_{i,k}$ can be normalized and serve as importance scores for the trees. However, there is often the case that there are multiple trees with the same weight, which counteracts the argument of a small number of models that can be easily inspected by a human. To mitigate this issue, we further differentiate the trees by calculating a second score, which indicates how "good" each model is for the nearest neighbors of each point. For example, a model $\mathcal{T}_i$ might have a large weight $u_{k,i}$ for a specific point $k$, because it might be in the top $L$ trees of the nearest neighbors, but its ranking, as defined in Section \ref{subsec:model_weighting}, might not be close to 1 for any of the neighbors. To address this, a larger score is given to trees that are also highly ranked across the nearest neighbors, and not just happen to be in the top $L$ models. 

More formally, for each point $k$ and tree $i$, $i=1, \dots, N$ two scores $S_{k,i,1}$ and $S_{k,i,2}$ are defined:
\begin{equation}
        S_{k.i,1} = \sum_{\boldsymbol{m} \in \mathcal{M}_k} \sum_{l=1}^{L} \mathbb{1} \{ \tau_{\boldsymbol{m}, (l)}  = \mathcal{T}_i\},
    \end{equation}
\begin{equation}
    S_{k.i,2} = - \sum_{\boldsymbol{m} \in \mathcal{M}_k} \sum_{l=1}^{L} l \cdot  \mathbb{1} \{ \tau_{\boldsymbol{m}, (l)}  = \mathcal{T}_i\}.
\end{equation}

$S_{k,i,1}$ is essentially equivalent to $u_{k,i}$, whereas $S_{k,i,2}$ corresponds to the negative sum of the rankings of the tree. Clearly, a tree is more important when the sum of the rankings is low, so the negative sign is added to make the two scores monotone in the same direction; namely the higher the score the better for both cases. After calculating these scores, they are normalized in the [0,1] range, and then the final score $S_{k,i}$ is calculated as:
\begin{equation}
    S_{k,i} = \frac{\bar S_{k,i,1} + \bar S_{k,i,2}}{2},
\end{equation}

\noindent where $\bar{S}_{k,i,1}$, $\bar{S}_{k,i,2}$ are the normalized scores $S_{k,i,1}$ and $S_{k,i,2}$, respectively. The final score contains information about how much a tree contributes to the final prediction for a specific sample, as well as how good it is at successfully classifying points from the training set that are close to the point of interest. For this reason, the trees with the highest scores can be utilized as reliable proxies to explain the final model's prediction in a personalized way. 

To illustrate this approach, we provide two examples from two different datasets we employ for our experiments in the Results Section \ref{subsec:res_model_weighting}.

\section{Results}

To demonstrate the performance and flexibility of our method, as well as its functionalities, we present a variety of different experiments and the corresponding results. In Section \ref{subsec:comparison} our Enhanced Random Forest approach is compared with other tree models, both with and without hyperparameter tuning. \textcolor{black} {It is noted that for the comparison with other models, all the extensions described in Section \ref{subsec:enhanced_rf} are considered. Sections \ref{subsec:res_sample_feature_importance} and \ref{subsec:res_model_weighting} illustrate the effect of the two main contributions (incorporating sample importance in the training and model weighting in the ensembling) separately; Section \ref{subsec:res_sample_feature_importance} discusses the improvements resulting from applying sample weights, non-uniform selection of samples for each tree, and feature selection at tree level, and also presents the benefits of sample and feature cleaning, whereas Section \ref{subsec:res_model_weighting} provides illustrative examples on how the tree weighting approach enables partial interpretability recovery and discusses how the performance of Enhanced Random Forests is either unaffected or improved by this approach.}

\subsection{Comparison with other models} \label{subsec:comparison}

The first group of experiments aims to show that Enhanced Random Forests significantly outperform vanilla Random Forests, and are better or comparable to other tree models. To establish that, the results in two cases are shown: both when their hyperparameters are tuned for each dataset, and when just the default hyperparameters across the datasets are selected and applied. 
In the first case, the best performance of the models for each dataset is studied, whereas the second case reveals, apart from the performance itself, how much finetuning is necessary to achieve satisfactory results. \textcolor{black} {The binary classification datasets we employed are presented in Table \ref{tab:datasets}, with 13 of them being publicly available.} The number of features is calculated after the one-hot encoding conversion of categorical variables to numerical. 

\begin{table}[h!]
\centering
\begin{tabular}{|c|c|c|c|}
\hline
Dataset Name & \# Instances & \# Features & Source/Reference \\
\hline
adult & 47,621 & 109 &  \cite{misc_adult_2} \\
aids & 15,000 & 23 & \cite{niaid_actg175_2001} \\
credit & 30,000 & 24 &  \cite{misc_default_of_credit_card_clients_350} \\
credit card & 36,457 & 59 & \cite{rikdifos_credit_card_2020} \\
deposit & 45,211 & 45 & \cite{misc_bank_marketing_222} \\
diabetes & 5,000 & 35 & - \\
gamma & 19,020 & 11 & \cite{misc_magic_gamma_telescope_159} \\
heart & 51,476 & 24 & \cite{lupague_integrated_2023} \\
hotel & 34,298 & 211 & \cite{antonio2019hotel} \\
loan & 25,535 & 29 & \cite{nikhil1e9_loan_default_2023} \\
online & 39,644 & 59 & \cite{misc_online_news_popularity_332} \\
shopping & 12,330 & 29 & \cite{misc_online_shoppers_purchasing_intention_dataset_468}\\
stroke & 4,909 & 22 & \cite{fedesoriano_stroke_prediction_2021} \\
spleen & 35,954 & 44 & - \\
wine & 6,497 & 12 & \cite{misc_wine_quality_186} \\
\hline
\end{tabular}
\caption{Summary of datasets used in the experiments.}
\label{tab:datasets}
\end{table}

\subsubsection{Results with hyperparameter tuning} \label{subsubsec:res_with_tuning}

The models we experimented with are boosting methods (XGBoost, Adaboost), vanilla Random Forests and CARTs. 

For all ensembles, we consider the same number of estimators (200). The hyperparameters considered for each model are:
\begin{itemize}
    \item \textbf{XGBoost}: We experimented with different maximum depth of each learner (3-6) and different learning rates in the range [0.01, 0.4].  
    \item \textbf{Random Forest}: We experimented with different values for the maximum depth (5,6).
    \item \textbf{CART}: We again experimented with different maximum depth values (5,6).
    \item \textbf{Enhanced Random Forest}: For our method, we had to tune multiple hyperparameters. These include whether the samples will be adaptively weighted, whether they will be selected with a non-uniform distribution, and how the features are selected in the trees. In the case of sample weighting or non-uniform probabilistic sample selection, we also need to specify a specific learning rate, as described in Section \ref{subsec:sample_weighting}. The best parameters per dataset can be found in the Appendix Section \ref{app_sec:opt_param}. We also employed model weighting in all cases.
    \item \textbf{Adaboost}: We experimented with depth values in the range 3-6 and different learning rates in the range [0.01, 1.0].
\end{itemize}

Our out-of-sample results are reported on a randomly selected hold-out test set, which comprises 15\% of the initial dataset. The hyperparameters are selected for each dataset separately, using 5-fold cross-validation. The values that give the best results are selected based on the average performance on the validation sets and then the average performance of the 5 models is reported on the test set. The metric we choose to report is AUC, since some of the datasets are imbalanced, making AUC a more suitable metric compared to accuracy. 

\begin{table}[!ht]
\setlength\extrarowheight{1.9pt}
\centering
\begin{tabular}{|c|c|c|c|c|c|}
\hline


Dataset & CART & Random Forest & \begin{tabular}{@{}c@{}} Enhanced  \\ Random Forest  \end{tabular} & XGBoost & Adaboost \\ \hline
adult & 0.8962 & 0.9039 & 0.9269 & \textbf{0.9291} & 0.9232 \\
aids & 0.6708 & \textbf{0.6965} & 0.6956 & 0.6939 & 0.6979 \\
credit & 0.7536 & 0.7754 & 0.7821 & \textbf{0.7825} & 0.7820 \\ 
credit card & 0.5743 & 0.6278 & \textbf{0.7038} & 0.6910 & 0.6449\\ 
deposit & 0.8576 & 0.9082 & 0.9274 & \textbf{0.9332} & 0.9268 \\ 
diabetes & 0.8660 & 0.8927 & 0.9187 & \textbf{0.9218} & 0.9131 \\ 
gamma & 0.8724 & 0.9031 & 0.9291 & \textbf{0.9334} & 0.9245 \\ 
heart  & 0.7967 & 0.8168 & 0.8201 & \textbf{0.8203} & 0.8173 \\ 
hotel & 0.8699 & 0.8989 & 0.9358 & \textbf{0.9434} & 0.9357 \\ 
loan & 0.7234 & 0.7670 & \textbf{0.7674} & 0.7651 & 0.7647 \\ 
online & 0.6894 & 0.7243 & 0.7347 & \textbf{0.7370} & 0.7337 \\ 
shopping & 0.9154 & 0.9203 & \textbf{0.9284} & 0.9257 & 0.9283 \\ 
stroke  & 0.7716 & \textbf{0.8475} & 0.8455 & 0.8379 & 0.8168 \\ 
spleen & 0.8616 & 0.9026 & 0.9167 & \textbf{0.9205} & 0.9189 \\ 
wine  & 0.8310 & 0.8622 & 0.9094 & 0.9123 & \textbf{0.9193} \\ 
\hline
\textbf{Average} & 0.7967 & 0.8298 & 0.8494 & \textbf{0.8500} & 0.8431 \\
\hline
\textbf{p-value} &  0.0002$^{***}$ & 0.0012$^{**}$ & - & 0.3028 & 0.0738  \\
\hline
\end{tabular}
\caption{Average AUC of all models we consider across the different train-validation splits (cross-validation). We used 15 different binary classification datasets for evaluation of the models.}
\label{tab:all_models_finetune}
\end{table}

From the results presented in Table \ref{tab:all_models_finetune}, we observe that our approach outperforms the CART and classic Random Forest models, with high statistical significance, as indicated by the adjusted p-values, calculated using the Wilcoxon test. It also performs comparably to boosting methods (XGBoost, AdaBoost), in a consistent fashion across the different datasets. This is evident from the calculated p-value as well; with high probability, XGBoost outperforming our method is due to randomness, although it performs slightly better in most datasets.

\subsubsection{Results without hyperparameter tuning}

As mentioned above, it is not only important to evaluate the best performance, but how much finetuning is required to achieve these results is also crucial. For this reason, the different models were trained and evaluated, using the default parameters proposed by the package implementations. For each type of model, we only select the maximum depth value that results in the best average AUC across all datasets. More specifically:
\begin{itemize}
    \item \textbf{XGBoost}: The default learning rate is 0.3. We select a maximum depth of 3, since this was the best across the datasets for the default learning rate.
    \item \textbf{Random Forest}: We use all default parameters, and a maximum depth of 6.
    \item \textbf{CART}: Again, we use all default parameters and a maximum depth of 6.
    \item \textbf{Enhanced Random Forest}: We keep the feature selection process per node, as in typical Random Forests, and we utilize adaptive sample weighting and non-uniform probabilistic selection of samples. We select the default learning rate to be 0.2, since, although it does not result in the best performance across the datasets, it provides a satisfactory trade-off between performance and speed. This approach of selecting the default hyperparameters follows the same rules applied for the selection of default parameters in available package implementations of the other models.
    \item \textbf{Adaboost}: The default learning rate is 1. We select a maximum depth of 6, since this was the best across the datasets, for the default learning rate. 
\end{itemize}

\begin{table}[!ht]
\setlength\extrarowheight{1.9pt}
\centering
\begin{tabular}{|c|c|c|c|c|c|}
\hline

Dataset & CART & Random Forest & \begin{tabular}{@{}c@{}} Enhanced  \\ Random Forest  \end{tabular} & XGBoost & Adaboost \\ \hline
adult & 0.8823 & 0.9039 & 0.9195 & \textbf{0.9295} & 0.9211 \\
aids & 0.6708 & 0.6965 & \textbf{0.6966} & 0.6768 & 0.6839 \\
credit & 0.7536 & 0.7754 & \textbf{0.7813} & 0.7770 & 0.7729 \\
credit card & 0.5676 & 0.6278 & \textbf{0.7038} & 0.6506 & 0.5912 \\
deposit & 0.8457 & 0.9082 & 0.9274 & \textbf{0.9336} & 0.9230 \\
diabetes & 0.8660 & 0.8927 & \textbf{0.9197} & 0.9150 & 0.9047 \\
gamma & 0.8634 & 0.9031 & 0.9291 & \textbf{0.9294} & 0.9206 \\
heart & 0.7951 & 0.8168 & \textbf{0.8188} & 0.8097 & 0.8075 \\
hotel & 0.8554 & 0.8989 & 0.9358 & \textbf{0.9396} & 0.9201 \\
loan & 0.7234 & 0.7670 & \textbf{0.7695} & 0.7468 & 0.7501 \\
online & 0.6835 & 0.7243 & \textbf{0.7329} & 0.7293 & 0.7214 \\
shopping & 0.9154 & 0.9203 & \textbf{0.9281} & 0.9155 & 0.9184 \\
stroke & 0.7716 & \textbf{0.8448} & 0.8403 & 0.7714 & 0.7864 \\
spleen & 0.8621 & 0.9026 & \textbf{0.9180} & 0.9128 & 0.9072 \\
wine & 0.8230 & 0.8623 & \textbf{0.9044} & 0.8960 & 0.8767 \\
\hline
\textbf{Average} & 0.7919 & 0.8296 & \textbf{0.8483} & 0.8355 & 0.8270 \\
\hline
\textbf{p-value} &  0.0002$^{***}$ & 0.0006$^{***}$ & - & 0.0256$^{*}$ & 0.0002$^{***}$  \\
\hline
\end{tabular}
\caption{Average AUC of all models we considered, without extensive finetuning. We used 15 different binary classification datasets for evaluation of the models.}
\label{tab:all_models_no_finetune}
\end{table}

The results presented in Table \ref{tab:all_models_no_finetune} demonstrate that our approach outperforms all the other tree methods with high statistical significance in the cases of CART, vanilla Random Forest and Adaboost, and with lower, but still considerable significance in the case of XGBoost. We observe that, compared to the finetuned version, Enhanced Random Forests do not have a considerable performance deterioration, unlike boosting methods. This indicates that our method is more likely to give better out-of-sample results using the default parameters, compared to other methods, and thus requires less finetuning. This indicates that our approach is more robust across datasets and the best hyperparameters do not deviate much from application to application. For the rest of the sections, the focus is given on exploring the effect of each of the proposed extensions on the Random Forest algorithm.

\subsection{Results on Sample and Feature Importance}\label{subsec:res_sample_feature_importance}
\textcolor{black}{This section discusses findings regarding the extensions related to Sample and Feature importance, as presented in Section \ref{subsec:sample_weighting}. The goal is to examine how these extensions function in practice, in addition to the overall performance improvements discussed in Section \ref{subsec:comparison}, and to offer insights into how each of them contributes to these improvements.}

\subsubsection{Effect of Enhanced Random Forest extensions} \label{subsec:effect_of_ext}

It is important to evaluate how each of our proposed additions perform separately, or in different combinations, for the various datasets in the Enhanced Random Forest framework. Since our proposed framework consists of multiple extensions, the exact combination for each dataset can be explored. Table \ref{tab:effect_of_hyperparameters} illustrates the performance of our method for different hyperparameter configurations. These include whether the random feature selection happens for each tree holistically \textcolor{black}{(FT)} or specifically for each split, whether sample weighting should be applied \textcolor{black}{(SW)}, or whether the selection of samples should occur following a non-uniform probability distribution \textcolor{black}{(SP)}, as described in Algorithm \ref{alg:training_with_weights}. The model weighting approach is not included in this case, to solely focus on the effect of the aforementioned extensions. 

\begin{table}[ht]
\centering
\begin{tabular}{lcccccccc}
\hline
 & Initial & FT & SW & SP & FT-SW & FT-SP & SW-SP & FT-SW-SP \\
\hline
adult & 0.9041 & 0.9046 & 0.9188 & 0.9186 & 0.9247 & 0.9260 & 0.9189 & \textbf{0.9267} \\
aids & 0.6965 & \textbf{0.6988} & 0.6969 & 0.6967 & 0.6987 & \textbf{0.6988} & 0.6973 & 0.6982 \\
credit & 0.7753 & 0.7712 & 0.7808 & 0.7809 & 0.7804 & 0.7804 & \textbf{0.7813} & 0.7808 \\
credit\_card & 0.6290 & 0.6353 & 0.6596 & 0.6566 & 0.6560 & 0.6471 & \textbf{0.6633} & 0.6501 \\
deposit & 0.9092 & 0.9053 & 0.9227 & 0.9225 & 0.9230 & 0.9204 & \textbf{0.9266} & 0.9219 \\
diabetes & 0.8931 & 0.8790 & \textbf{0.9189} & 0.9188 & 0.9163 & 0.9140 & 0.9186 & 0.9139 \\
gamma & 0.9040 & 0.9082 & 0.9248 & 0.9255 & 0.9255 & 0.9242 & \textbf{0.9279} & 0.9264 \\
heart & 0.8164 & 0.8120 & 0.8198 & \textbf{0.8201} & 0.8166 & 0.8156 & 0.8187 & 0.8158 \\
hotel & 0.8977 & 0.8552 & 0.9289 & 0.9246 & 0.9181 & 0.9192 & \textbf{0.9313} & 0.9261 \\
loan & 0.7660 & 0.7583 & 0.7666 & 0.7671 & 0.7560 & 0.7558 & \textbf{0.7696} & 0.7578 \\
online & 0.7240 & 0.7255 & \textbf{0.7340} & \textbf{0.7340}  & 0.7335 & 0.7305 & 0.7333 & 0.7312 \\
shopping & 0.9194 & 0.9000 & 0.9261 & 0.9261 & 0.9157 & 0.9134 & \textbf{0.9272} & 0.9138 \\
stroke & 0.8420 & 0.8372 & 0.8430 & 0.8432 & 0.8360 & 0.8372 & \textbf{0.8434} & 0.8372 \\
spleen & 0.9024 & 0.9027 & 0.9160 & 0.9164 & 0.9128 & 0.9132 & \textbf{0.9181} & 0.9124 \\
wine & 0.8637 & 0.8677 & 0.8950 & 0.8963 & 0.8991 & 0.8992 & 0.8979 & \textbf{0.8997} \\
\hline
\end{tabular}
\caption{Out-of-sample average AUC across the different datasets, for different hyperparameter configurations. Dictionary for the columns:
\newline FT: features being selected randomly for each tree and not for each separate split, as in typical Random Forests. 
\newline SW: sample weights.
\newline SP: samples selected with a specific probability at each tree.}
\label{tab:effect_of_hyperparameters}
\end{table}

It is evident that for the majority of the datasets, sample weighting, combined with definition of specific probabilities for the sample selection, is the most beneficial approach. More specifically, in 11 out of the 15 datasets, although both approaches separately improve upon the vanilla Random Forest approach, their combination gives better results than each of the two methods individually. Selecting features at the tree level instead of the split level appears to be beneficial in 3 out of 15 datasets, especially in the adult dataset, where it leads to a performance improvement of around 1\%. 

These results highlight the effect of our proposed extension the different datasets, revealing the considerable improvement they offer compared to the vanilla Random Forest implementation. \textcolor{black}{Most importantly, it underlines the consistent improvement of the combination of sample weighting and sampling with custom probabilities for the majority of the datasets, indicating their primary contribution in the performance enhancement discussed in Section \ref{subsec:comparison}.}

\subsubsection{Sample and Feature cleaning} \label{subsec:res_sample_feature_cleaning}
In this Section, the main goal is to explore how our framework can be leveraged to clean the dataset, both in terms of features and in terms of samples. Following the approach discussed in Section \ref{subsec:sample_feature_cleaning}, the optimal parameters for each dataset (\ref{app_sec:opt_param}) are used and our approach is run iteratively, as described in Algorithm \ref{alg:sample_feature_cleaning}. After each iteration, we inspect:
\begin{itemize}
    \item \textbf{Sample importance}: In both the sample weighting and non-uniform probabilistic selection of samples, a weight is assigned to each sample. The samples that have low scores are discarded, as described in Section \ref{subsec:sample_feature_cleaning}.
    \item \textbf{Feature importance}: The feature importance is calculated for each feature, based on its contribution on the tree splits. The samples whose importance is less than 10\% of the highest feature importance are discarded. 
\end{itemize}

The maximum number of iterations is set to 70; also, an early stopping criterion is adopted, where if the AUC on the validation set is more than 1\% lower than the maximum AUC obtained up to this iteration, the process terminates. These choices were made after experimentation, and can, of course, be adapted, based on the users' needs.

The goal is to explore how much the dataset can be reduced, so that the out-of-sample AUC drop is less than 0.2\% (0.002) compared to the initial AUC, obtained with the full dataset. In the experiment presented in Table \ref{tab:auc_feature_sample_reduction}, sample and feature selection are performed simultaneously. The results of individual sample and feature selection are presented in the Appendix Section \ref{app_sec:feat_sample_cleaning}. The relative AUC difference refers to the difference between the AUC after and before the sample and feature selection.

\begin{table}[ht]
\centering
\begin{tabular}{|c|c|c|c|}
\hline
Dataset & \begin{tabular}{@{}c@{}}  Relative AUC \\ difference (\%) \end{tabular} & \begin{tabular}{@{}c@{}} Feature \\ reduction(\%)  \end{tabular}  &  \begin{tabular}{@{}c@{}} Sample size \\ reduction(\%) \end{tabular}\\
\hline
adult & 0.02 & 54.26 & 22.54 \\
aids & 0.07 & 26.36 & 6.55 \\
credit & -0.15 & 2.61 & 2.90 \\
credit card & 0.40 & 5.17 & 0.05 \\
deposit & -0.11 & 43.64 & 3.10 \\
diabetes & 0.02 & 23.64 & 39.72 \\
gamma & -0.14 & 0.00 & 36.36 \\
heart & -0.07 & 23.48 & 0.24 \\
hotel & 0.08 & 63.62 & 43.31 \\
loan & 0.18 & 42.14 & 1.25 \\
online & -0.18 & 18.28 & 14.63 \\
shopping & -0.10 & 42.14 & 4.70 \\
stroke & 0.10 & 16.19 & 0.21 \\
spleen & -0.17 & 43.72 & 1.42 \\
wine & 0.08 & 0.00 & 14.53 \\
\hline
\end{tabular}
\caption{Average Relative AUC difference(\%), Feature reduction(\%), and Sample size reduction(\%) for various datasets.}
\label{tab:auc_feature_sample_reduction}
\end{table}

We observe that for 8 out of 15 datasets, cleaning the dataset is actually beneficial, even when the dataset size is significantly reduced. In the rest of the cases, a much smaller dataset results in a minor performance deterioration. On average, we can reduce features by 26.55\% and samples by 11.01\% with a performance sacrifice of less than 0.2\% in AUC.

It is worth noting that some datasets benefit from both feature and sample cleaning, where others only from one of the two, or from none. Overall, this cleaning process can contribute in keeping the most important parts of the dataset and reduce computational time when running the model, as well as memory requirements. It also offers interesting insights into which samples and features are actually crucial for obtaining a reliable model.

\textcolor{black}{These results highlight the additional benefit of introducing the concept of sample importance to the model, as a way to clean large datasets without sacrificing performance. It also provides an automatic indication of how redundant the information is in the observational data, and potentially also removes outliers that hurt the model's performance.}

\subsection{Results on Model weighting} \label{subsec:res_model_weighting}
In this section, the goal is to explore the second extension of our approach compared to the vanilla random forest, and especially how assigning weights to each tree of the ensemble, as described in Section \ref{subsec:model_weighting} compares to treating each tree equally. 

\subsubsection{Interpretability} \label{subsubsec:interpretability}
\textcolor{black}{The main contribution of this approach is its leveraging to create an importance score for each tree and then select the most significant ones for each sample, as described in Section \ref{subsec:interpretability}. To illustrate the benefit of this method, we present examples from two different datasets we employed. This visualization clearly demonstrates the edge bagging can have compared to boosting in terms of interpretability, since the trees that comprise the boosting ensemble are not predictors for the classification problem.}

\paragraph{Example 1: Spleen dataset}
The spleen dataset contains information from 35,954 patients, who have suffered a spleen injury. The features of the patients include demographics, like age and sex, and medical information, like pulse oximetry, pulse rate, respiratory rate, the severity of their injuries etc. Depending on the severity of their diagnosis and their clinical and demographic features, the patients either receive simple observation (treatment a) or get treated with splenectomy (complete removal of the spleen - treatment b) or angioembolization (minimally invasive technique - treatment c). The outcome we want to predict in this problem is patient mortality (binary outcome). 

For this reason, a simple Random Forest model is trained, and then used to make predictions for each patient in the test set. A typical Random Forest is selected for simplicity in this example, and to illustrate that the model weighting idea can be applied to a vanilla Random Forest as well, without the need for our extended framework. For each patient, a probability of belonging to class 1 (death) is the output of the model, and in order to make the final prediction, a cut-off threshold between the two classes is specified, as described in Section \ref{subsec:sample_weighting}. The optimal threshold is selected using the Youden's J statistic. For this specific example, the selected threshold $t$ is 0.0699, meaning that if the output probability from the model is greater than $t$ then the patient is classified as 1, and 0 otherwise. We note that $t$ is much smaller than the typical 0.5 threshold value; this is attributed to the highly imbalanced nature of the dataset.

For the purposes of this example, we are focusing on a specific patient $k$ from the test set, who eventually died. Some of the features of this patient, used by the trees, are presented in Table \ref{tab:spleen_selected_features}. Comparing with the median values of the training set, we observe that patient $k$ has a considerably low pulse rate, low pulse oximetry and total GCS, so some of these variables are expected to be decisive factors in the highest scoring trees. The ensemble correctly classifies this patient as 1, since the output probability after applying the tree weights is $p_k = 0.1726 > 0.0699=t$. We note that adding tree weights offered a small improvement in out-of-sample AUC on the full test set from 0.8922 to 0.8955. Using the method described in Section \ref{subsec:interpretability}, a score for each tree considered in the ensemble is calculated. The top 3 models with the highest scores, 0.9528, 0.9344 and 0.9337 respectively, are selected. These are the most contributing models to the final prediction.

\begin{table}[]
\centering
\begin{tabular}{>{\raggedright}p{4cm} >{\raggedleft\arraybackslash}p{3cm} >{\raggedleft\arraybackslash}p{3cm}}
\toprule
\textbf{Feature} & \textbf{Values for Patient $k$} & \textbf{Median Values} \\
\midrule
sex & 1 & 1 \\
age & 24 & 38 \\
pulse rate & 115 & 94 \\
respiratory rate & 16 & 19 \\
pulse oximetry & 72 & 98 \\
total gcs & 3 & 15 \\
cc\_cva (stroke) & 0 & 0 \\
cc\_smoking & 0 & 0 \\
cc\_cirrhosis & 0 & 0 \\
liver\_inj & 0 & 0 \\
treatment\_b (splenectomy) & 0 & 0 \\
\bottomrule
\end{tabular}
\caption{Selected features and their values for patient $k$ and the corresponding median values of the training set.}
\label{tab:spleen_selected_features}
\end{table}

\begin{figure}
  \centering
   \includegraphics[width=\linewidth]{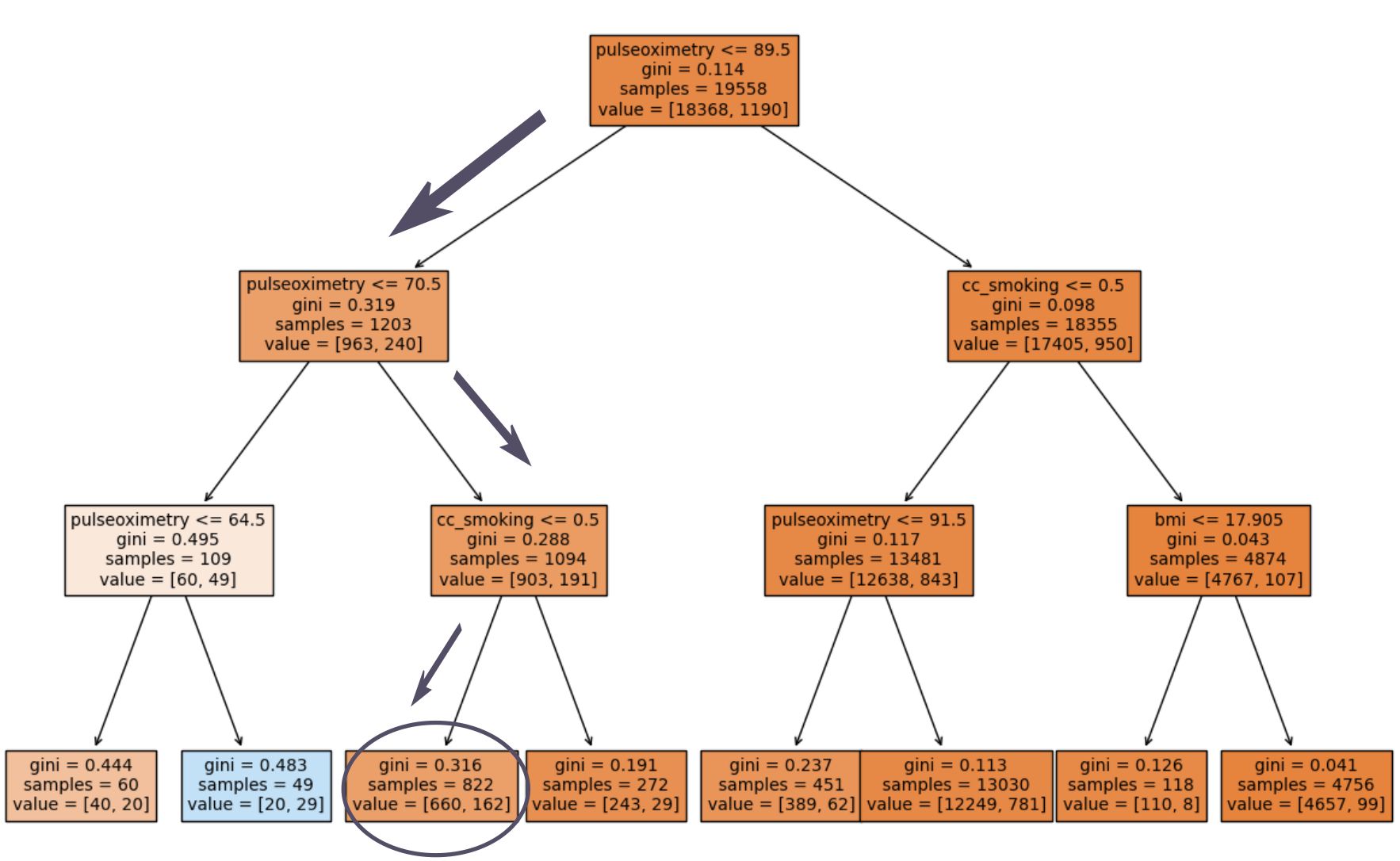}

   \caption{Highest-scoring tree for Patient $k$.}
   \label{fig:spleen_1}
\end{figure}

In Figure \ref{fig:spleen_1} the highest-scoring tree is presented. The rest can be found in the Appendix Section \ref{app_sec:trees}. We observe that the model picks on the low pulse oximetry of the patient compared to the median as the most determinant factor. Overall, by investigating the most important models we conclude that the most significant variables that led to the prediction of patient $k$ are their pulse oximetry, pulse rate, respiratory rate and smoking status. Though the calculated scores are a proxy, since the final prediction depends on other tree models too, it can narrow down the most contributing trees to a small number that can be manually inspected. In this way, a clinician could, for example, investigate and track down the most prominent factors that led to the final prediction for each patient independently, which offers a personalized and interpretable treatment of each data sample.

\paragraph{Example 2: Shopping dataset}

This dataset consists of 12,330 rows (website sessions) that correspond to different users in 1-year periods, and the goal is to predict the customer's purchasing intention. The available features contain information about the type of pages that were visited in each session, how much time was spent on these pages, metrics calculated by Google Analytics that refer to the different pages in the e-commerce site, as well as the time of the session in the year. We again train a typical Random Forest, and we apply weights to the trees. In this case, this approach results in a considerable increase in the model's performance on the test set, with the AUC increasing from 0.9019 to 0.9183.

We are focusing on a specific session $k$ that resulted in purchase. This was successfully predicted by our ensemble model, that gave an output probability of purchase equal to 0.8288. Some basic features of this session are presented in Table \ref{tab:shopping_selected_features}. The three first features correspond to number of administrative, informational and product related pages that were visited in the session. The PageValues variable is the average value of pages visited during the session, as calculated by Google Analytics, and the last variable indicates whether the user is new or returning. We observe that, comparing to the median values of the features in the training set, the user spent a considerable amount of time on product related pages, they are returning customers and the Page Value of the sites they visited is very high, which indicates the high profit associated with these pages, an observation which is expected to be reflected on the highest-scoring trees. 

\begin{table}[]
\centering
\begin{tabular}{>{\raggedright}p{4cm} >{\raggedleft\arraybackslash}p{3cm} >{\raggedleft\arraybackslash}p{3cm}}
\toprule
\textbf{Feature} & \textbf{Values for Session $k$} & \textbf{Median Values} \\
\midrule
Administrative & 0 & 1\\
Informational & 0 & 0\\
ProductRelated & 18 & 24 \\
PageValues & 75.79 & 0 \\
VisitorType\_Returning\_Visitor & 1 &1\\
\bottomrule
\end{tabular}
\caption{Selected features and their values for session $k$ and the corresponding median values of the training set.}
\label{tab:shopping_selected_features}
\end{table}

\begin{figure}
  \centering
   \includegraphics[width=\linewidth]{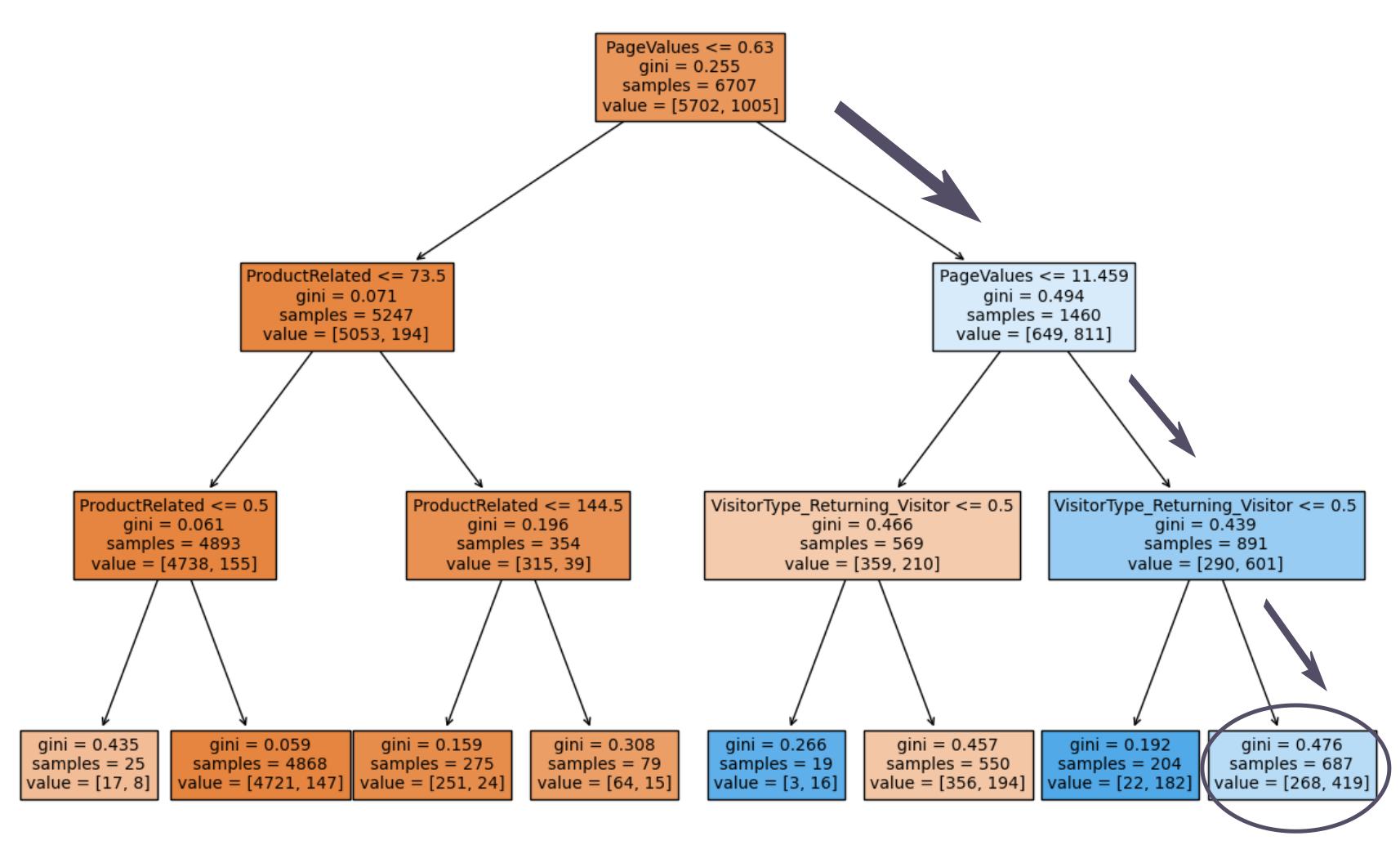}

   \caption{Highest-scoring tree for Session $k$.}
   \label{fig:shopping_1}
\end{figure}

Unlike the previous example, for this sample the second highest normalized score was 0.83, so the final prediction is heavily defined by the highest scoring tree, displayed in Figure \ref{fig:shopping_1}. As expected, the PageValues feature is the most important in making the final prediction. The result is highly interpretable and matches our intuition on the prediction. 

\subsubsection{Model weighting}
\begin{figure}
  \centering
   \includegraphics[width=\linewidth]{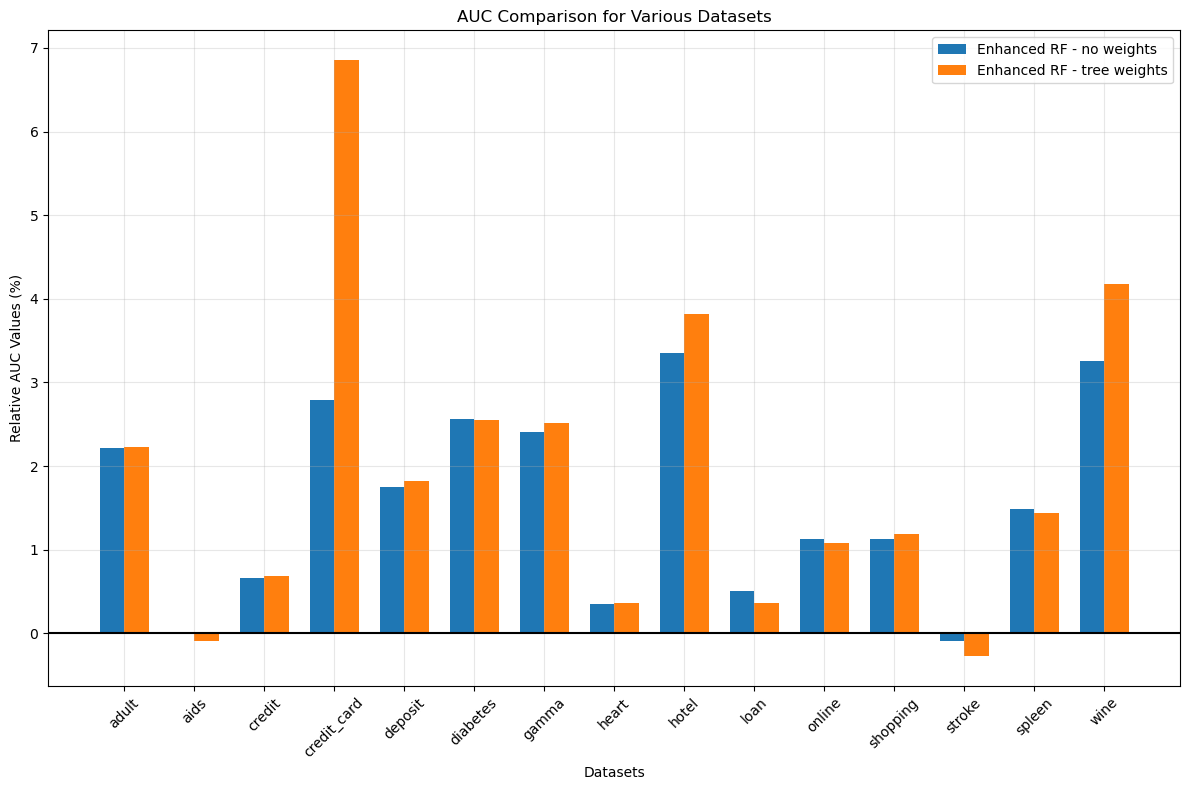}

   \caption{Relative improvement of Enhanced Random Forests with and without tree weights. }
   \label{fig:barplot_van_enh_mw}
\end{figure}

Apart from the interpretability aspect, the effect of model weights on the performance of the ensemble is explored. For this experiment, the best hyperparameter combination for each dataset is selected, and then compared to the typical Random Forest. The best hyperparameters are the ones selected using cross-validation, as in Section \ref{subsubsec:res_with_tuning}. In order to focus on the relative scores between the vanilla and the Enhanced Random Forest, the difference of average AUC between the Enhanced Random Forest with and without tree weights compared to the vanilla Random Forest for each dataset is evaluated, as presented in Figure \ref{fig:barplot_van_enh_mw}. The scores presented are $\text{AUC}_{E,NW} - \text{AUC}_V$ and $\text{AUC}_{E,MW} - \text{AUC}_V$, where $\text{AUC}_{E,NW}$ is the average AUC of the Enhanced Random Forest without model weights, $\text{AUC}_{E,MW}$ is the average AUC of the Enhanced Random Forest with model weights, and $\text{AUC}_{V}$ is the average AUC of the vanilla Random Forest implementation.

We observe that the majority of the datasets benefit significantly from applying our framework to the typical Random Forest implementation. The tree weights provide comparable or better performance in most cases. \textcolor{black} {More specifically, the Enhanced Random Forest framework is beneficial for 13 out of 15 datasets. Including model weights in the framework offers an extra improvement in 6 of the datasets, with a maximum performance boost of around 4\% in one of the datasets. Even in the cases where the performance is not improved, the results are comparable, and the user can benefit from the interpretability option offered, as shown in Section \ref{subsubsec:interpretability}.}

\begin{figure}
  \centering
   \includegraphics[width=\linewidth]{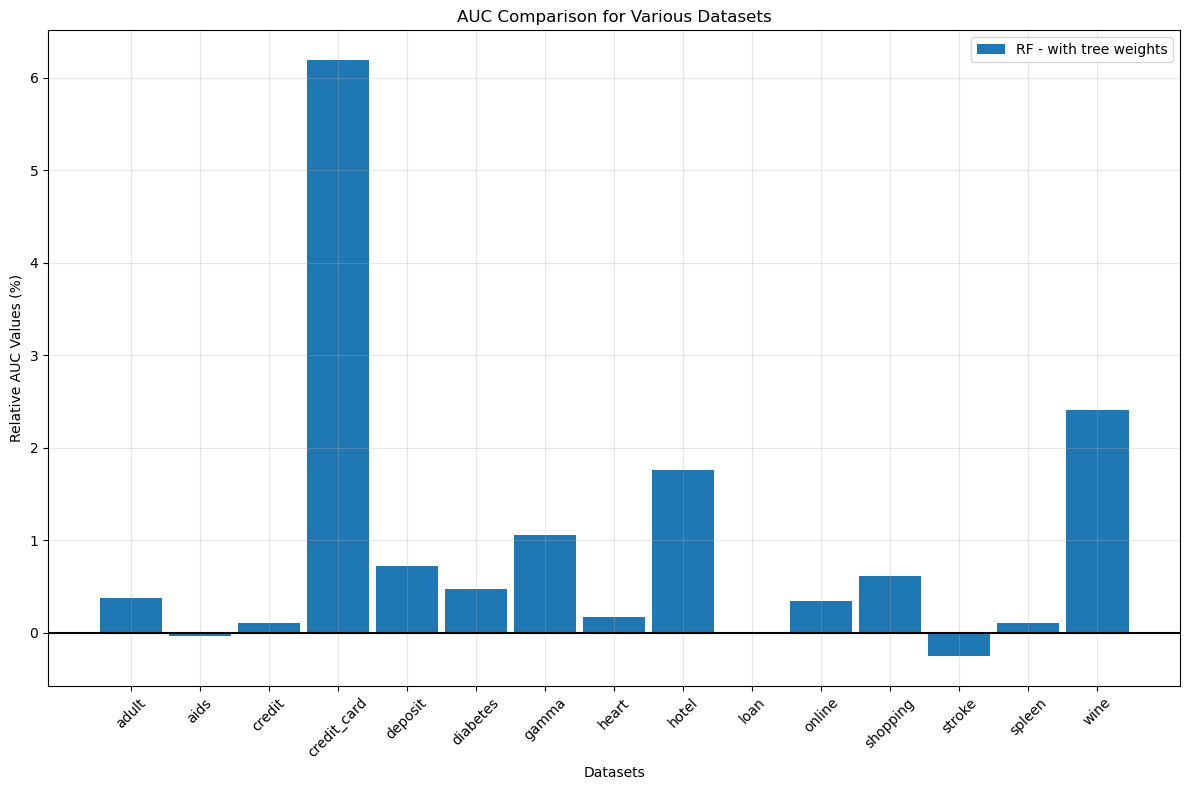}

   \caption{Relative improvement of Random Forests with tree weights. }
   \label{fig:barplot_van_mw}
\end{figure}

When the weights are applied to the vanilla Random Forest, without using the rest of the extensions discussed in Section \ref{subsec:sample_weighting}, the performance enhancement is more evident, as demonstrated in Figure \ref{fig:barplot_van_mw}. In this plot, the relative improvement of a Random Forest with tree weights compared to equal weights for all trees is presented. \textcolor{black} {The performance enhancement is apparent in 13 out of 15 datasets; by only applying model weights, average AUC on the test set increases by more than 1\% in 4 datasets, and more than 2\% in 2 datasets. }

We observe that in most cases, adding weights to the trees using our approach is beneficial. 
\textcolor{black}{Overall, model weighting enables the introduction of interpretability in our methodology, without sacrificing performance, highlighting the edge of bagging methods compared to boosting in the explainability section. The interpretable nature of individual Random Forest learners is leveraged in our approach to get an insight into which factors contribute to the prediction in a more concrete way, compared to the usual feature importance or Shapley values exploration.}


\section{Conclusions}
In this work, we present an extension of vanilla Random Forests, called Enhanced Random Forests, that facilitates more functionalities and presents a more flexible and general tool for bagging approaches. It improves both the performance and the interpretability of Random Forests, empirically suggesting that they can be as powerful as boosting approaches for binary classification in terms of performance, and also more explainable, a feature that is in general critically missing from ensemble algorithms. \textcolor{black} {This indicates that bagging methods can benefit from extensions that aim to reduce the element of randomness, perform comparably to boosting methods, and, combined with the interpretable nature of their individual learners, be considered a more appealing method than boosting for binary classification problems, which are the most commonly encountered problems in predictive machine learning.}

A natural extension of our method would be to adjust the framework for multi-classification problems. The model weighting approach can be similar; in terms of the sample weighting, the update of the weights can again be proportional to the misclassification error, but this should be defined differently. For instance, one approach could be to define an optimal classification threshold for each class in an one-vs-all setting and update the weight based on the difference between the output probability of the sample belonging to its ground-truth class and the threshold defined for this class. 

Focusing on our work, incorporating sample importance as part of the training of Random Forests, by favoring the hardest training points, proves to empirically boost their performance and make it comparable to the state-of-the-art boosting methods. We also show that without hyperparameter tuning for each dataset individually, Enhanced Random Forests outperform all other tree methods, indicating that our approach can be successfully utilized with the default parameters we suggest, without the need for excessive tuning. 

Sample weights provide a notion of sample importance, and along with calculated feature importance scores, we show that our framework can also be used to clean the training dataset, both in terms of features and samples, resulting in more manageable dataset sizes, without sacrificing performance. We show that in multiple datasets, cleaning the dataset is actually beneficial, implying that there might often be excessive noise in the data. On average, more than 25\% of the features and 10\% of the samples can be dropped across datasets, with minimum performance deterioration.  

In addition, using weights for each tree of the ensemble in a personalized way further improves the performance in several datasets, but also enables users to partially recover interpretability, since it provides them with the trees that mostly contribute in the final prediction for each sample. This feature is critical for multiple applications, for example in healthcare settings, where it is crucial for users to have an insight into the most important factors that are more essential for the final output of the model. This is a significant differentiation of our approach compared to boosting methods, which cannot benefit from this idea, since the learners that are used are fitted on residuals and are, thus, not intuitive. 

\textcolor{black}{Boosting methods, being fast and accurate, are considered one of the state-of-the-art approaches for classification tasks when working with tabular data; however we show that Enhanced Random Forests can be a valuable, or even stronger alternative to boosting methods for binary classification problems. They empirically provide similar performance, they do not require extensive hyperparameter tuning, can be used to remove the least helpful samples and features from the training set, and offer a partial interpretability option, which is significant for a plethora of applications and is crucially absent from boosting methods.}

\section*{Funding}
This work was conducted by the authors while they were affiliated with the Massachusetts Institute of Technology.

\section*{Ethics declarations}
\subsection*{Conflicts of interest/Competing interests}
Not applicable.
\subsection*{Ethics approval}
Not applicable.
\subsection*{Consent to participate}
Not applicable.
\subsection*{Consent for publication}
Not applicable.
\subsection*{Availability of data and material}
All datasets used are publicly available, except for two (diabetes, spleen), which contain medical data from collaborating hospitals.

\subsection*{Code availability}
The code implementing the experiments is available upon request. 

\section*{Authors' contributions}
Dimitris Bertsimas motivated, supported and supervised the presented work and edited the paper. 
Vasiliki Stoumpou developed the presented algorithms, implemented the pipeline and the code, ran the experiments and wrote the paper.

\bibliographystyle{alpha}
\bibliography{sn-bibliography}

\newcommand{\etalchar}[1]{$^{#1}$}
\begin{thebibliography}{MRGOGP12}

\bibitem[BAB21]{barratt2021optimal}
Shane Barratt, Guillermo Angeris, and Stephen Boyd.
\newblock Optimal representative sample weighting.
\newblock {\em Statistics and Computing}, 31:1--14, 2021.

\bibitem[BK96]{misc_adult_2}
Barry Becker and Ronny Kohavi.
\newblock {Adult}.
\newblock UCI Machine Learning Repository, 1996.
\newblock {DOI}: https://doi.org/10.24432/C5XW20.

\bibitem[BLCW09]{bengio2009curriculum}
Yoshua Bengio, J{\'e}r{\^o}me Louradour, Ronan Collobert, and Jason Weston.
\newblock Curriculum learning.
\newblock In {\em Proceedings of the 26th annual international conference on
  machine learning}, pages 41--48, 2009.

\bibitem[Bra07]{bramer2007avoiding}
Max Bramer.
\newblock Avoiding overfitting of decision trees.
\newblock {\em Principles of data mining}, pages 119--134, 2007.

\bibitem[Bre96]{breiman1996bagging}
Leo Breiman.
\newblock Bagging predictors.
\newblock {\em Machine learning}, 24:123--140, 1996.

\bibitem[Bre01]{breiman2001random}
Leo Breiman.
\newblock Random forests.
\newblock {\em Machine learning}, 45:5--32, 2001.

\bibitem[Bre17]{breiman2017classification}
Leo Breiman.
\newblock {\em Classification and regression trees}.
\newblock Routledge, 2017.

\bibitem[CG16]{chen2016xgboost}
Tianqi Chen and Carlos Guestrin.
\newblock Xgboost: A scalable tree boosting system.
\newblock In {\em Proceedings of the 22nd acm sigkdd international conference
  on knowledge discovery and data mining}, pages 785--794, 2016.

\bibitem[CJL{\etalchar{+}}19]{cui2019class}
Yin Cui, Menglin Jia, Tsung-Yi Lin, Yang Song, and Serge Belongie.
\newblock Class-balanced loss based on effective number of samples.
\newblock In {\em Proceedings of the IEEE/CVF conference on computer vision and
  pattern recognition}, pages 9268--9277, 2019.

\bibitem[CY18]{misc_online_shoppers_purchasing_intention_dataset_468}
Sakar C. and Kastro Yomi.
\newblock {Online Shoppers Purchasing Intention Dataset}.
\newblock UCI Machine Learning Repository, 2018.
\newblock {DOI}: https://doi.org/10.24432/C5F88Q.

\bibitem[CYZ23]{chen2023optimal}
Xinyu Chen, Dalei Yu, and Xinyu Zhang.
\newblock Optimal weighted random forests.
\newblock {\em arXiv preprint arXiv:2305.10042}, 2023.

\bibitem[EMLB16]{Elvira2016HereticalMI}
V{\'i}ctor Elvira, Luca Martino, David Luengo, and M{\'o}nica~F. Bugallo.
\newblock Heretical multiple importance sampling.
\newblock {\em IEEE Signal Processing Letters}, 23:1474--1478, 2016.

\bibitem[fed21]{fedesoriano_stroke_prediction_2021}
fedesoriano.
\newblock Stroke prediction dataset.
\newblock
  \url{https://www.kaggle.com/datasets/fedesoriano/stroke-prediction-dataset},
  2021.
\newblock Accessed: 2024-06-04.

\bibitem[FKP15]{misc_online_news_popularity_332}
Cortez~Paulo Fernandes~Kelwin, Vinagre~Pedro and Sernadela Pedro.
\newblock {Online News Popularity}.
\newblock UCI Machine Learning Repository, 2015.
\newblock {DOI}: https://doi.org/10.24432/C5NS3V.

\bibitem[FS97]{freund1997decision}
Yoav Freund and Robert~E Schapire.
\newblock A decision-theoretic generalization of on-line learning and an
  application to boosting.
\newblock {\em Journal of computer and system sciences}, 55(1):119--139, 1997.

\bibitem[IC16]{misc_default_of_credit_card_clients_350}
Yeh I-Cheng.
\newblock {Default of Credit Card Clients}.
\newblock UCI Machine Learning Repository, 2016.
\newblock {DOI}: https://doi.org/10.24432/C55S3H.

\bibitem[JMZ{\etalchar{+}}15]{jiang2015self}
Lu~Jiang, Deyu Meng, Qian Zhao, Shiguang Shan, and Alexander Hauptmann.
\newblock Self-paced curriculum learning.
\newblock In {\em Proceedings of the AAAI Conference on Artificial
  Intelligence}, volume~29, 2015.

\bibitem[KKMA11]{kim2011weight}
Hyunjoong Kim, Hyeuk Kim, Hojin Moon, and Hongshik Ahn.
\newblock A weight-adjusted voting algorithm for ensembles of classifiers.
\newblock {\em Journal of the Korean Statistical Society}, 40(4):437--449,
  2011.

\bibitem[KM53]{kahn1953methods}
Herman Kahn and Andy~W Marshall.
\newblock Methods of reducing sample size in monte carlo computations.
\newblock {\em Journal of the Operations Research Society of America},
  1(5):263--278, 1953.

\bibitem[KPK10]{kumar2010self}
M~Kumar, Benjamin Packer, and Daphne Koller.
\newblock Self-paced learning for latent variable models.
\newblock {\em Advances in neural information processing systems}, 23, 2010.

\bibitem[LB02]{li2002instability}
Ruey-Hsia Li and Geneva~G Belford.
\newblock Instability of decision tree classification algorithms.
\newblock In {\em Proceedings of the eighth ACM SIGKDD international conference
  on Knowledge discovery and data mining}, pages 570--575, 2002.

\bibitem[LGG{\etalchar{+}}17]{lin2017focal}
Tsung-Yi Lin, Priya Goyal, Ross Girshick, Kaiming He, and Piotr Doll{\'a}r.
\newblock Focal loss for dense object detection.
\newblock In {\em Proceedings of the IEEE international conference on computer
  vision}, pages 2980--2988, 2017.

\bibitem[Lun17]{lundberg2017unified}
Scott Lundberg.
\newblock A unified approach to interpreting model predictions.
\newblock {\em arXiv preprint arXiv:1705.07874}, 2017.

\bibitem[LWDD10]{li2010trees}
Hong~Bo Li, Wei Wang, Hong~Wei Ding, and Jin Dong.
\newblock Trees weighting random forest method for classifying high-dimensional
  noisy data.
\newblock In {\em 2010 IEEE 7th international conference on e-business
  engineering}, pages 160--163. IEEE, 2010.

\bibitem[LZ17]{liu2017variable}
Yiyi Liu and Hongyu Zhao.
\newblock Variable importance-weighted random forests.
\newblock {\em Quantitative Biology}, 5(4):338--351, 2017.

\bibitem[MRGOGP12]{maudes2012random}
Jes{\'u}s Maudes, Juan~J Rodr{\'\i}guez, C{\'e}sar Garc{\'\i}a-Osorio, and
  Nicol{\'a}s Garc{\'\i}a-Pedrajas.
\newblock Random feature weights for decision tree ensemble construction.
\newblock {\em Information Fusion}, 13(1):20--30, 2012.

\bibitem[MSP12]{misc_bank_marketing_222}
Rita~P. Moro~S. and Cortez P.
\newblock {Bank Marketing}.
\newblock UCI Machine Learning Repository, 2012.
\newblock {DOI}: https://doi.org/10.24432/C5K306.

\bibitem[{Nat}01]{niaid_actg175_2001}
{National Institute of Allergy and Infectious Diseases (NIAID)}.
\newblock A randomized, double-blind phase ii/iii trial of monotherapy vs.
  combination therapy with nucleoside analogs in hiv-infected persons with cd4
  cells of 200-500/mm3.
\newblock ClinicalTrials.gov Identifier: NCT00000625, 2001.
\newblock Other Study ID Numbers: ACTG 175, 11150 (Registry Identifier: DAIDS
  ES Registry Number). First Posted: August 31, 2001. Last Update Posted:
  November 2, 2021. Last Verified: October 2021.

\bibitem[NdAAL19]{antonio2019hotel}
Antonio Nuno, de~Almeida~Ana, and Nunes Luis.
\newblock Hotel booking demand datasets.
\newblock {\em Data in brief}, 22:41--49, 2019.

\bibitem[nik23]{nikhil1e9_loan_default_2023}
nikhil1e9.
\newblock Loan default prediction dataset.
\newblock \url{https://www.kaggle.com/datasets/nikhil1e9/loan-default},
  September 2023.
\newblock Accessed: 2024-06-04.

\bibitem[PPBV19]{Paananen2019PushingTL}
Topi Paananen, Juho Piironen, Paul-Christian B{\"u}rkner, and Aki Vehtari.
\newblock Pushing the limits of importance sampling through iterative moment
  matching.
\newblock {\em arXiv: Computation}, 2019.

\bibitem[R.07]{misc_magic_gamma_telescope_159}
Bock R.
\newblock {MAGIC Gamma Telescope}.
\newblock UCI Machine Learning Repository, 2007.
\newblock {DOI}: https://doi.org/10.24432/C52C8B.

\bibitem[rik20]{rikdifos_credit_card_2020}
rikdifos.
\newblock Credit card approval prediction.
\newblock
  \url{https://www.kaggle.com/datasets/rikdifos/credit-card-approval-prediction},
  2020.
\newblock Accessed: 2024-06-04.

\bibitem[RRAM23]{lupague_integrated_2023}
Lupague R.M.J.M., Mabborang R.C., Bansil A.G., and Lupague M.M.
\newblock Integrated machine learning model for comprehensive heart disease
  risk assessment based on multi-dimensional health factors.
\newblock {\em European Journal of Computer Science and Information
  Technology}, 11(3):44--58, 2023.

\bibitem[R{\v{S}}04]{robnik2004improving}
Marko Robnik-{\v{S}}ikonja.
\newblock Improving random forests.
\newblock In {\em European conference on machine learning}, pages 359--370.
  Springer, 2004.

\bibitem[RZYU18]{ren2018learning}
Mengye Ren, Wenyuan Zeng, Bin Yang, and Raquel Urtasun.
\newblock Learning to reweight examples for robust deep learning.
\newblock In {\em International conference on machine learning}, pages
  4334--4343. PMLR, 2018.

\bibitem[S{\etalchar{+}}99]{schapire1999brief}
Robert~E Schapire et~al.
\newblock A brief introduction to boosting.
\newblock In {\em Ijcai}, volume~99, pages 1401--1406. Citeseer, 1999.

\bibitem[SBS{\etalchar{+}}21]{santiago2021low}
Carlos Santiago, Catarina Barata, Michele Sasdelli, Gustavo Carneiro, and
  Jacinto~C Nascimento.
\newblock Low: Training deep neural networks by learning optimal sample
  weights.
\newblock {\em Pattern recognition}, 110:107585, 2021.

\bibitem[SH21]{shahhosseini2021improved}
Mohsen Shahhosseini and Guiping Hu.
\newblock Improved weighted random forest for classification problems.
\newblock In {\em Progress in Intelligent Decision Science: Proceeding of IDS
  2020}, pages 42--56. Springer, 2021.

\bibitem[SXY{\etalchar{+}}19]{shu2019meta}
Jun Shu, Qi~Xie, Lixuan Yi, Qian Zhao, Sanping Zhou, Zongben Xu, and Deyu Meng.
\newblock Meta-weight-net: Learning an explicit mapping for sample weighting.
\newblock {\em Advances in neural information processing systems}, 32, 2019.

\bibitem[TJ09]{misc_wine_quality_186}
Cortez Paulo Cerdeira A. Almeida F.~Matos T. and Reis J.
\newblock {Wine Quality}.
\newblock UCI Machine Learning Repository, 2009.
\newblock {DOI}: https://doi.org/10.24432/C56S3T.

\bibitem[WFB13]{winham2013weighted}
Stacey~J Winham, Robert~R Freimuth, and Joanna~M Biernacka.
\newblock A weighted random forests approach to improve predictive performance.
\newblock {\em Statistical Analysis and Data Mining: The ASA Data Science
  Journal}, 6(6):496--505, 2013.

\bibitem[XLL18]{xuan2018refined}
Shiyang Xuan, Guanjun Liu, and Zhenchuan Li.
\newblock Refined weighted random forest and its application to credit card
  fraud detection.
\newblock In {\em Computational Data and Social Networks: 7th International
  Conference, CSoNet 2018, Shanghai, China, December 18--20, 2018, Proceedings
  7}, pages 343--355. Springer, 2018.

\end{thebibliography}
\newpage
\section{Appendix}

\subsection{Optimal hyperparameters} \label{app_sec:opt_param}

\begin{table}[h!]
\centering
\begin{tabular}{>{\raggedright}p{3cm} >{\raggedright}p{2cm} >{\raggedright}p{2cm} >{\raggedright}p{2cm} >{\raggedright\arraybackslash}p{1.5cm} >{\raggedright\arraybackslash}p{1.5cm}}
\toprule
\textbf{Dataset} & \textbf{Sample probabilities} & \textbf{Sample weights} & \textbf{Feature selection per tree} & \textbf{Max depth} & \textbf{LR} \\
\midrule
adult         & True  & True  & True & 6 & 0.20 \\
aids          & True  & False & False  & 5 & 0.05 \\
credit        & True  & True  & False  & 6 & 0.05 \\
credit card  & True  & False & True & 6 & 1.00 \\
deposit       & True  & True  & False  & 6 & 0.10 \\
diabetes      & True  & False & False  & 6 & 0.05 \\
gamma         & True  & True  & False  & 6 & 0.05 \\
heart         & True  & True  & False  & 6 & 0.05 \\
hotel         & True  & True  & False  & 6 & 0.20 \\
loan          & True  & True  & False  & 5 & 0.10 \\
online        & False & True  & False  & 6 & 0.10 \\
shopping      & True  & True  & False  & 5 & 0.05 \\
stroke        & True  & False & False  & 5 & 0.50 \\
spleen        & True  & True  & False  & 6 & 0.10 \\
wine          & True  & False & True & 6 & 0.10 \\
\bottomrule
\end{tabular}
\caption{Best Parameters for Various Datasets, selected using Cross Validation}
\label{tab:config_params}
\end{table}


\newpage
\subsection{Effects of Feature and Sample cleaning} \label{app_sec:feat_sample_cleaning}

\begin{table}[ht]
\centering
\begin{tabular}{|c|c|c|c|}
\hline
Dataset & \begin{tabular}{@{}c@{}}  Relative AUC \\ difference (\%) \end{tabular} & \begin{tabular}{@{}c@{}} Feature \\ reduction(\%)  \end{tabular}  &  \begin{tabular}{@{}c@{}} Sample size \\ reduction(\%) \end{tabular}\\\hline
adult & -0.03 & 63.89 & 0.00 \\
aids & 0.01 & 32.73 & 0.00 \\
credit & -0.07 & 1.74 & 0.00 \\
credit\_card & 0.51 & 5.86 & 0.00 \\
deposit & -0.14 & 49.55 & 0.00 \\
diabetes & 0.13 & 18.18 & 0.00 \\
gamma & 0.00 & 0.00 & 0.00 \\
heart & -0.19 & 25.22 & 0.00 \\
hotel & -0.08 & 67.62 & 0.00 \\
loan & -0.07 & 47.86 & 0.00 \\
online & 0.05 & 16.55 & 0.00 \\
shopping & -0.05 & 53.57 & 0.00 \\
stroke & 0.12 & 33.33 & 0.00 \\
spleen & -0.15 & 57.21 & 0.00 \\
wine & 0.00 & 0.00 & 0.00 \\
\hline
\end{tabular}
\caption{Average Relative AUC difference(\%), Feature reduction(\%), and Sample size reduction(\%) for various datasets, when we only perform feature selection.}
\label{app_tab:auc_feature_reduction}
\end{table}

\begin{table}[ht]
\centering
\begin{tabular}{|c|c|c|c|}
\hline
Dataset & \begin{tabular}{@{}c@{}}  Relative AUC \\ difference (\%) \end{tabular} & \begin{tabular}{@{}c@{}} Feature \\ reduction(\%)  \end{tabular}  &  \begin{tabular}{@{}c@{}} Sample size \\ reduction(\%) \end{tabular}\\
\hline
adult & -0.02 & 0.00 & 20.75 \\
aids & -0.08 & 0.00 & 6.84 \\
credit & -0.14 & 0.00 & 3.41 \\
credit\_card & 0.66 & 0.00 & 0.10 \\
deposit & -0.16 & 0.00 & 6.96 \\
diabetes & -0.06 & 0.00 & 40.66 \\
gamma & -0.14 & 0.00 & 36.36 \\
heart & -0.18 & 0.00 & 1.37 \\
hotel & 0.02 & 0.00 & 39.74 \\
loan & 0.06 & 0.00 & 12.18 \\
online & -0.13 & 0.00 & 14.56 \\
shopping & -0.01 & 0.00 & 4.60 \\
stroke & 0.64 & 0.00 & 0.23 \\
spleen & -0.17 & 0.00 & 2.69 \\
wine & 0.08 & 0.00 & 14.53 \\
\hline
\end{tabular}
\caption{Average Relative AUC difference(\%), Feature reduction(\%), and Sample size reduction(\%) for various datasets, when we only perform sample selection.}
\label{app_tab:auc_sample_reduction}
\end{table}

\newpage

\subsection{Tree models examples for the spleen dataset} \label{app_sec:trees}

\begin{figure}[h!]
  \centering
   \includegraphics[width=\linewidth]{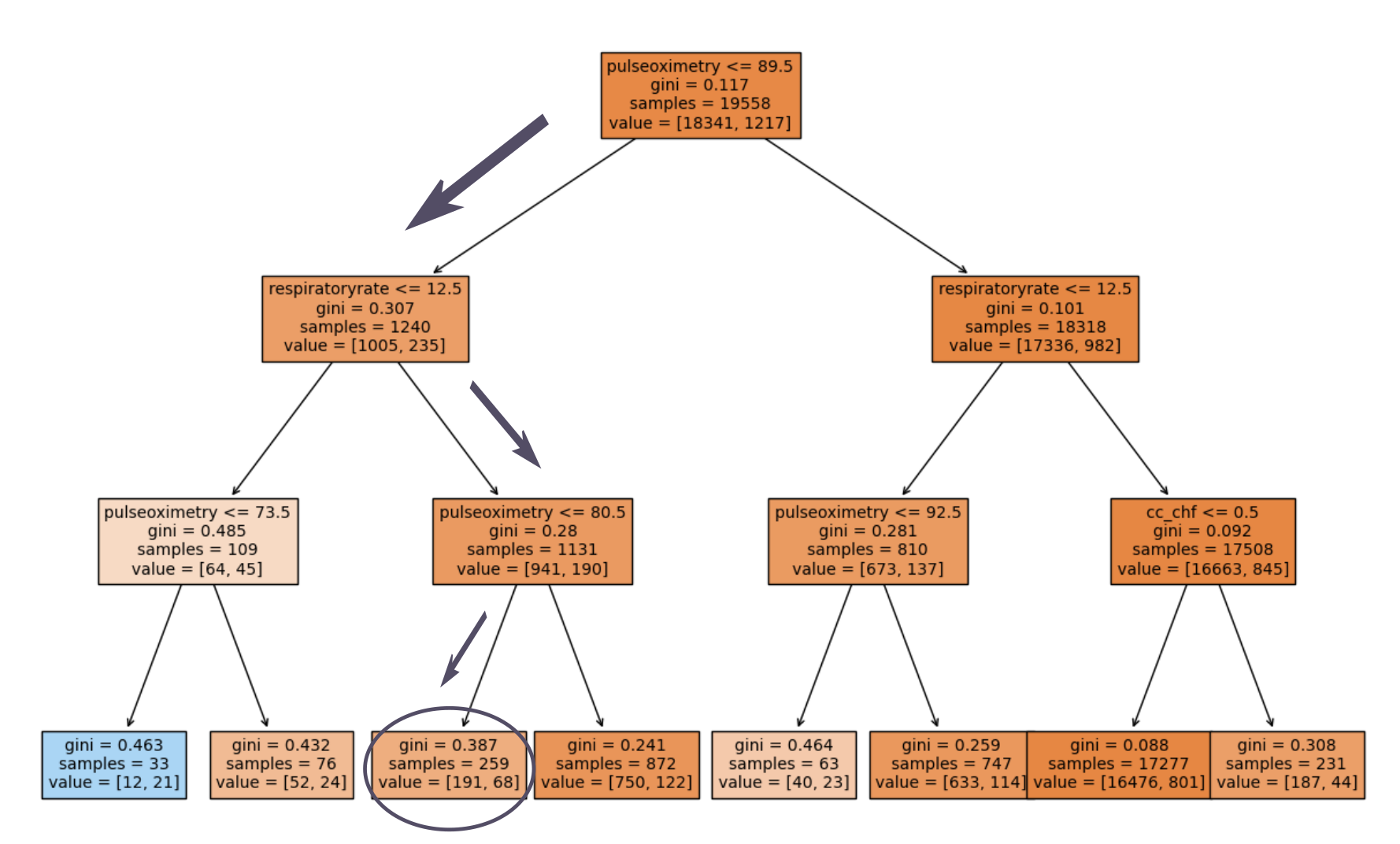}

   \caption{Second highest-scoring tree for Patient $k$.}
   \label{fig:spleen_2}
\end{figure}

\begin{figure}[h!]
  \centering
   \includegraphics[width=\linewidth]{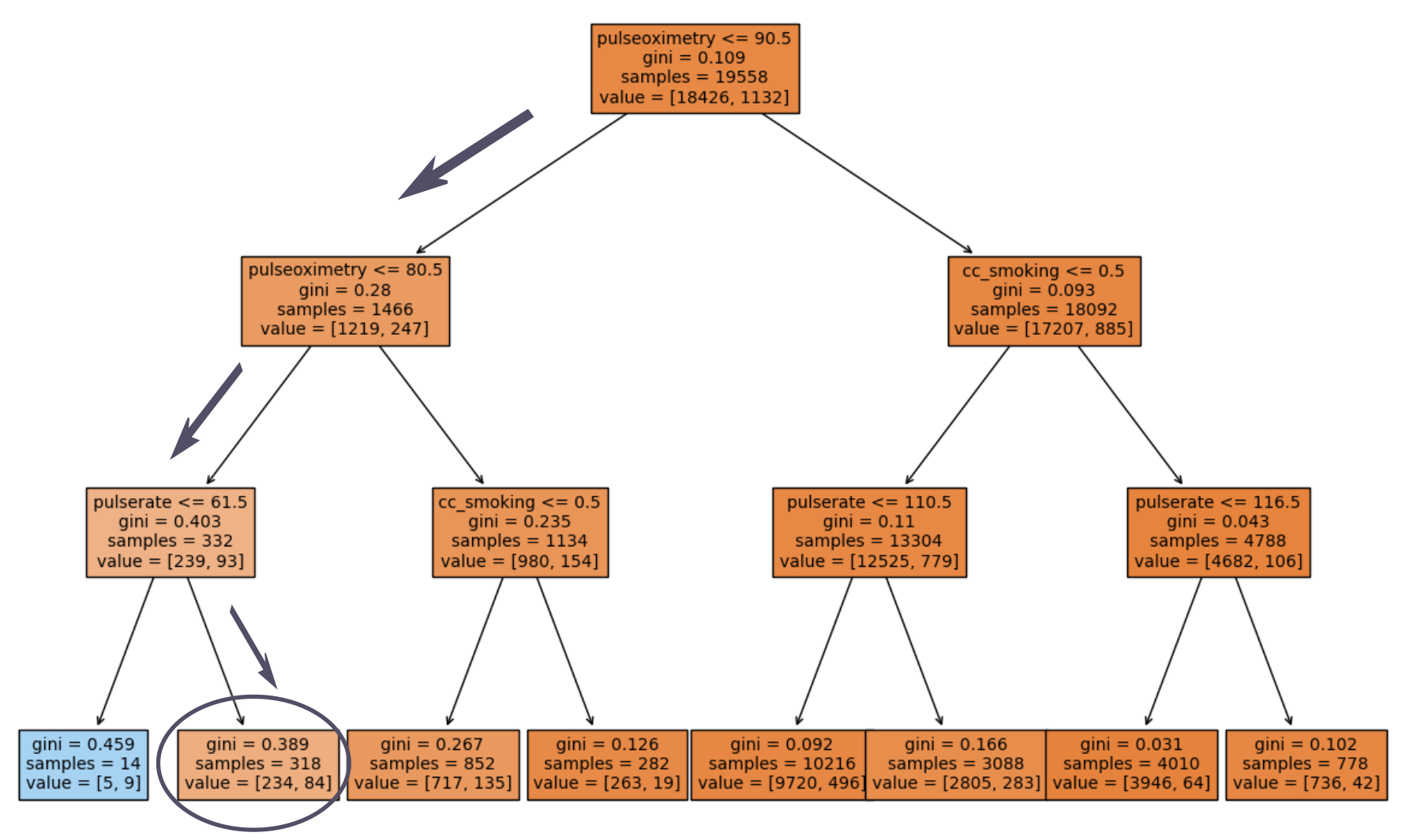}

   \caption{Third highest-scoring tree for Patient $k$.}
   \label{fig:spleen_3}
\end{figure}



\end{document}